%% file: iclr2026_conference.tex
\documentclass{article} 
\usepackage{iclr2026_conference,times}

\input{math_commands.tex}

\usepackage{hyperref}
\usepackage{url}

\usepackage{subcaption}
\usepackage[utf8]{inputenc} 
\usepackage[T1]{fontenc}    
\usepackage{hyperref}       
\usepackage{booktabs}       
\usepackage{amsfonts}       
\usepackage{bbm}            
\usepackage{nicefrac}       
\usepackage{microtype}      
\usepackage[table]{xcolor}         

\usepackage{amssymb}
\usepackage{mathtools}
\usepackage{enumitem}
\usepackage{amsthm}
\usepackage{amsmath, algorithm, algpseudocode}
\usepackage{tabularx}
\usepackage{multirow} 
\usepackage{pifont}
\usepackage{fontawesome5}
\usepackage{tcolorbox}
\usepackage{listings}
\usepackage{caption}
\usepackage{ragged2e}
\usepackage{marvosym}
\tcbuselibrary{listings,breakable,skins}

\definecolor{szcgreen}{rgb}{0.0, 0.5, 0.0}
\definecolor{szcblue}{rgb}{0.0, 0.53, 0.74}
\definecolor{szcred}{rgb}{0.7, 0.11, 0.11}
\definecolor{backcolour}{rgb}{0.95,0.95,0.95}

\input{marcos.tex}

\title{
TimeSearch-R: Adaptive Temporal Search for Long-Form Video Understanding via Self-Verification Reinforcement Learning
}

\author{Junwen Pan$^{1}$\thanks{Equal contribution. \quad $^{\dagger}$Project lead. \quad $^{\text{\Letter}}$Corresponding author.} \quad Qizhe Zhang$^{1,2\, *}$ \quad Rui Zhang$^{1}$ \quad Ming Lu$^{2\, \dagger}$ \\
\textbf{Xin Wan$^{1}$ \quad Yuan Zhang$^{1,2}$ \quad Chang Liu$^{1}$ \quad Qi She$^{1\, \text{\Letter}}$} \\
$^1$ ByteDance \quad $^2$ School of Computer Science, Peking University \\
\texttt{\{panjunwen,sheqi.roger\}@bytedance.com}
}

\iclrfinalcopy 
\begin{document}

\maketitle

\begin{abstract}
Temporal search aims to identify a minimal set of relevant frames from tens of thousands based on a given query, serving as a foundation for accurate long-form video understanding. 
Existing works attempt to progressively narrow the search space. 
However, these approaches typically rely on a hand-crafted search process, lacking end-to-end optimization for learning optimal search strategies. 
In this paper, we propose \textbf{TimeSearch-R}, which reformulates temporal search as interleaved text-video thinking, seamlessly integrating searching video clips into the reasoning process through reinforcement learning (RL). 
However, applying RL training methods, such as Group Relative Policy Optimization (GRPO), to video reasoning can result in unsupervised intermediate search decisions. 
This leads to insufficient exploration of the video content and inconsistent logical reasoning. 
To address these issues, we introduce GRPO with Completeness Self-Verification (GRPO-CSV), which gathers searched video frames from the interleaved reasoning process and utilizes the same policy model to verify the adequacy of searched frames, thereby improving the completeness of video reasoning.
Additionally, we construct datasets specifically designed for the SFT cold-start and RL training of GRPO-CSV, filtering out samples with weak temporal dependencies to enhance task difficulty and improve temporal search capabilities.
Extensive experiments demonstrate that TimeSearch-R achieves significant improvements on temporal search benchmarks such as Haystack-LVBench and Haystack-Ego4D, as well as long-form video understanding benchmarks like VideoMME and MLVU. 
Notably, TimeSearch-R establishes a new state-of-the-art on LongVideoBench with 4.1\% improvement over the base model Qwen2.5-VL and 2.0\% over the advanced video reasoning model Video-R1. \textit{Our code is available at \href{https://github.com/Time-Search/TimeSearch-R}{https://github.com/Time-Search/TimeSearch-R}.}

\end{abstract}

\section{Introduction}

Long-form video understanding requires models to navigate through tens of thousands of frames to identify the most relevant information for answering specific questions~\citep{videomme, MLVU, wu2024longvideobench}. 
Temporal search lies at the heart of making long-video understanding both accurate and interpretable~\citep{park2025lvnet, li2023discovering, ye2025rethinking}. 
In contrast to the human visual system, which conducts adaptive temporal search~\citep{yarbus1967eye, hayhoe2005eye}, current large video-language models (LVLMs) primarily rely on hand-crafted search strategies with static frame sampling~\citep{video-llava,bai2025qwen25vltechnicalreport, feng2025videor1}. 
Humans naturally alternate between broad scanning and targeted inspection, refining their focus iteratively based on intermediate findings~\citep{castelhano2009optimizing, henderson2017meaning}. 
In contrast, existing methods are limited to a fixed set of frames established before the reasoning process begins.
This design presents a fundamental contradiction: video reasoning is a dynamic process where temporal search interleaves with video reasoning; however, the video frames accessible to the model remain fixed from the outset, ultimately hindering effective reasoning.

\begin{figure}[t]
    \centering
    \includegraphics[width=1.0\textwidth]{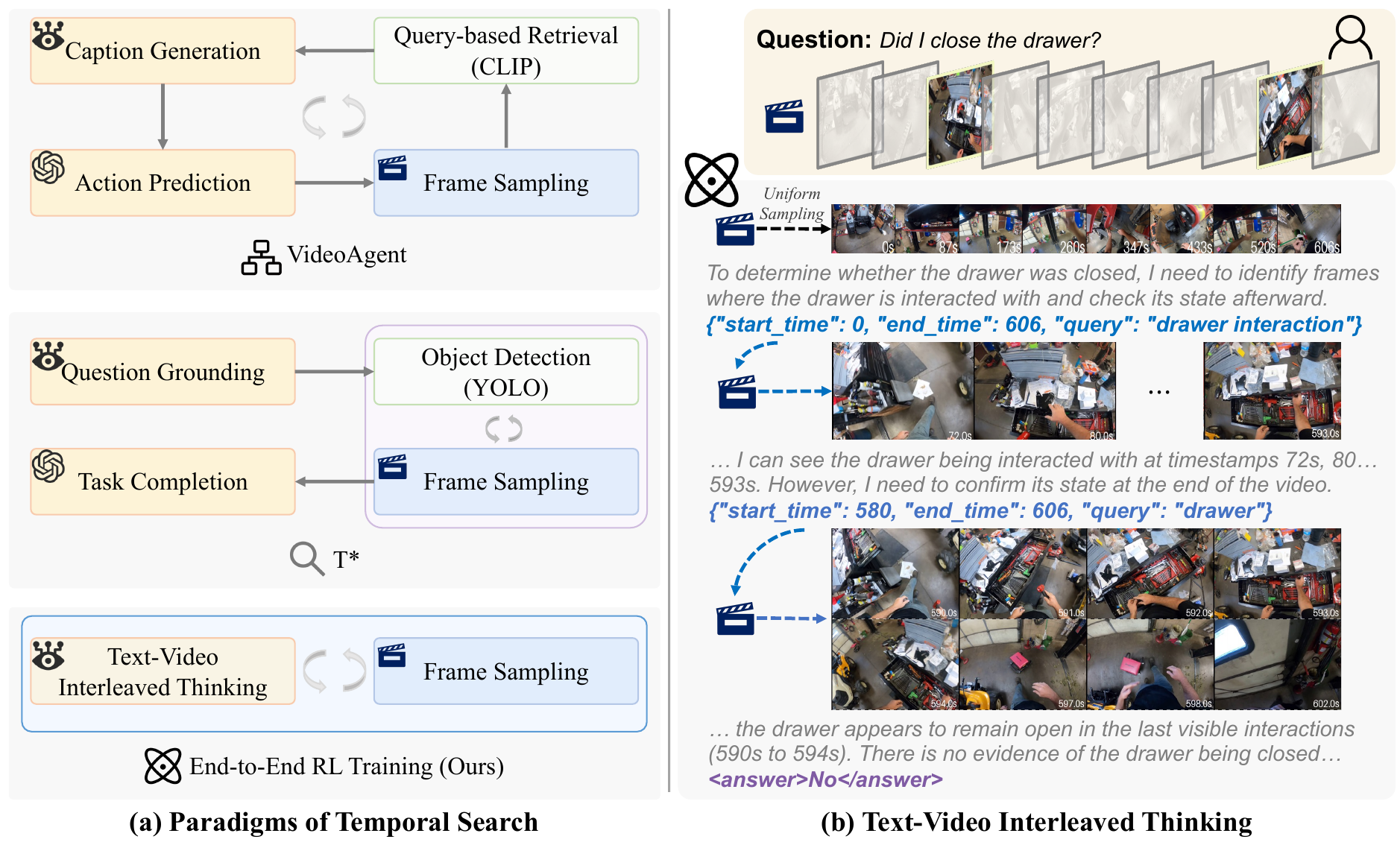}
    \vspace{-3mm}
    \caption{\textbf{(a) Different paradigms of temporal search.} Previous works such as VideoAgent~\citep{wang2024videoagent} and T*~\citep{ye2025rethinking} predominantly rely on handcrafted workflows, resulting in suboptimal strategies. Our approach adopts end-to-end reinforcement learning, enabling the model to learn optimal search strategies directly from data. \textbf{(b) Interleaved text-video thinking process.} We reformulate the temporal search task as an interleaved text-video thinking process, where the temporal search is seamlessly interleaved into the reasoning process.}
    \label{fig:teaser}
    \vspace{-3mm}
\end{figure}

Inspired by the gap between human cognition and model reasoning, recent studies have explored interactive video agents that attempt to bridge this divide through multi-turn temporal search, as illustrated in Figure~\ref{fig:teaser} (a). VideoAgent~\citep{wang2024videoagent} first employs a large language model (LLM) as the central agent, which iteratively calls tools like vision-language models (VLMs) and CLIP~\citep{clip} for frame captioning and retrieval, and then aggregates information in the textual modality to perform reasoning and predict answers. T*~\citep{ye2025rethinking} extends this paradigm by introducing an object-oriented spatial-temporal search. It first leverages a VLM to extract target objects from the question, then employs object detection models (e.g., YOLO~\citep{cheng2024yolo}) to identify keyframes containing these objects, and finally uses the retrieved frame set to complete the task. Moreover, strategies that introduce tree-structured search to improve efficiency have also been explored~\citep{wang2025videotree, li2025iterativezoomin, pan2025timesearch}. However, all of these approaches depend on manually designed workflows, which lead to suboptimal search strategies.

This motivates us to explore an end-to-end learning approach that discovers optimal temporal search strategies directly from data. In this work, we reformulate the temporal search task as an interleaved text-video thinking process, and propose \textbf{\textsc{TimeSearch-R}}, a model that learns to actively search for relevant temporal clips through reinforcement learning (RL). As shown in Figure~\ref{fig:teaser} (b), our model alternates between textual reasoning and temporal exploration, iteratively refining its understanding of the video. We refer to this dynamic process as \textbf{Thinking with Videos}—a paradigm where models gradually improves their comprehension by searching for relevant video content conditioned on intermediate reasoning states. This concept extends the recent advances in multimodal reasoning, \textit{Thinking with Images}~\citep{thinkingwithimages, hu2024visual, deepeyes}, to the long-video domain.

\begin{figure}[t]
    \centering
    \includegraphics[width=1.0\textwidth]{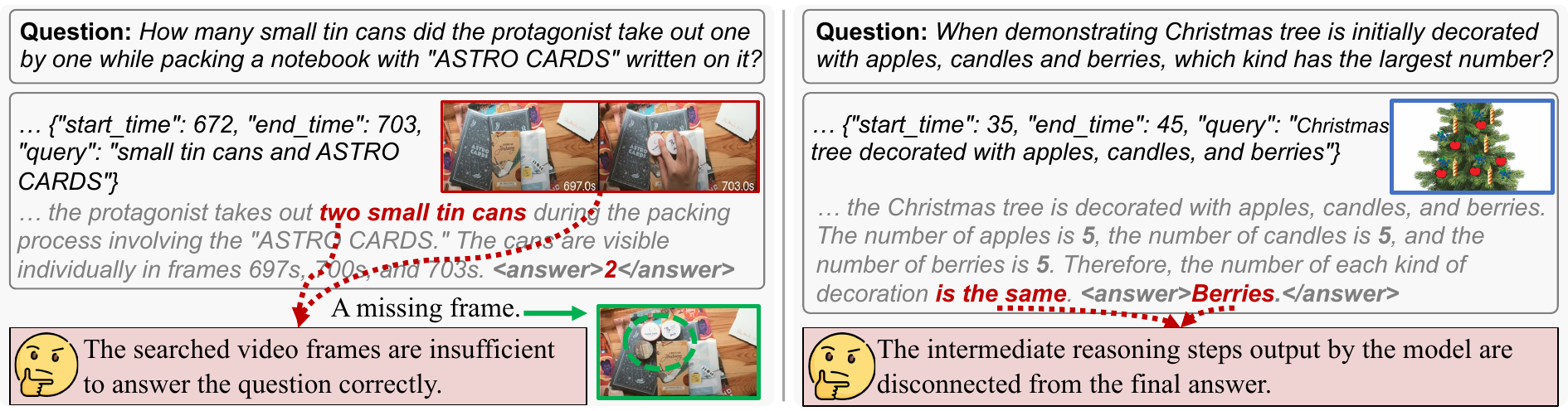}
    \caption{\textbf{Two failure modes with the original GRPO reward.} Left: \textbf{Insufficient temporal exploration.} The model misses critical frames required to correctly answer the question. Right: \textbf{Inconsistent logical reasoning.} The intermediate reasoning process contradicts the final answer.}
    \label{fig:failures}
\end{figure}

Although recent works have successfully applied RL algorithms like Group Relative Policy Optimization (GRPO)~\citep{deepseekai2025deepseekr1incentivizingreasoningcapability} to textual~\citep{jin2025search_r1} and spatial search~\citep{deepeyes}, temporal search in videos poses unique challenges. The original GRPO rewards only the final output while ignoring intermediate search decisions, leading to several failure modes illustrated in Figure~\ref{fig:failures}. The first mode, termed \textbf{insufficient temporal exploration}, arises because the final output reward provides no incentive for comprehensive exploration of video frames. LVLMs may arrive at correct answers through partial evidence or language bias without proper visual grounding~\citep{niu2021counterfactual}, missing critical frames required for reliable understanding. The second mode, termed \textbf{inconsistent logical reasoning}, emerges when models produce plausible thinking processes disconnected from the final answers, a phenomenon also observed in text-only reasoning~\citep{MeasuringFaithfulness}. These two failure modes hinder proper temporal search and diminish the benefits of video reasoning.

To address these challenges, we propose \textbf{Completeness Self-Verification (CSV)} as a supplement to the original GRPO algorithm, providing supervision over the intermediate steps of temporal search. GRPO-CSV tackles insufficient temporal exploration by ensuring the model to acquire sufficient visual evidence through self-verification, and promotes consistency between intermediate reasoning and the final answer by re-answering the question using the searched frames. Besides, we construct a high-quality video reasoning dataset to support GRPO-CSV training. Existing datasets contain a large number of trivial samples solvable through prue linguistic bias, as well as noisy samples that remain unsolvable even with extensive search, severely hindering progress in long-video reasoning. We implement a two-stage data filtering pipeline to curate high-quality samples tailored to the demands of video reasoning, ensuring that the model learns the correct process of temporal search.

We evaluate our TimeSearch-R on both temporal search and long-form video understanding tasks, demonstrating its superiority in long video reasoning. On temporal search tasks, TimeSearch-R improving the temporal F1 score on Haystack-LVBench by 5.6\% and the accuracy on Haystack-Ego4D by 8.5\%, compared to the previous state-of-the-art (SOTA) method. On long-form video understanding tasks, TimeSearch-R establishes new SOTA results with 4.1\% improvement over the base model Qwen2.5-VL and 2.0\% over the advanced reasoning model Video-R1 on LongVideoBench.

In summary, our main contributions are three-fold:
\begin{itemize}
    \item [1.] We propose the \textbf{TimeSearch-R} framework, which reformulates temporal search as interleaved text-video thinking and learns optimal search strategies directly from data.
    \item [2.] We introduce \textbf{GRPO-CSV}, a novel RL algorithm, which ensures sufficient and accurate video exploration by supervising the intermediate steps of temporal search. To support GRPO-CSV training, we also construct a high-quality video reasoning dataset via a two-stage filtering pipeline, enabling the model to learn correct temporal search processes.
    \item [3.] Extensive experiments demonstrate the superiority of our approach on both temporal search and long-form video understanding. Notably, TimeSearch-R establishes a new SOTA on LongVideoBench, outperforming the latest reasoning model Video-R1 by \textbf{2.0\%}.
\end{itemize}

\section{Methods}

In this section, we first reformulate the temporal search task as an interleaved text-video thinking process, enabling the model to learn optimal search strategies directly from data. To address the challenges of insufficient temporal exploration and inconsistent logical reasoning, we introduce GRPO-CSV as a novel RL algorithm for long videos, which ensures both sufficient and accurate video exploration by supervising intermediate steps of temporal search. Finally, we describe the model training process, including the construction of a high-quality long-video reasoning dataset.

\subsection{Task Formulation}
\label{sec:task_formulation}

\paragraph{Temporal Search within Thinking Process.} To learn optimal search strategies directly from data, we reformulate temporal search as a multi-turn thinking process interleaved with video clip retrieval. Given a video $V$ and a corresponding question $Q$, an initial preview $\tilde{V}$ is uniformly sampled from $V$ for subsequent reasoning. At each thinking step $k$, the policy model $\pi_{\theta}$ generates a textual reasoning $T_k$. If $T_k$ contains a \texttt{search} instruction, the video environment executes it according to frame timestamps, retrieving a clip $V_k \subseteq V$ that is appended to the ongoing chain of thought (CoT) as input for later steps. The interleaved text-video CoT at reasoning step $k$ is formalized as:
\begin{equation}
    C_k \,\triangleq\, \{\, (T_1,\, V_1),\, (T_2,\, V_2),\, \ldots,\, (T_k,\, V_k)\,\}.
    \label{cot_formulation}
\end{equation}
This interaction process repeats until the model emits the final answer $A$ or reaches the pre-defined reasoning budget. For further analysis, the entire reasoning chain can be decomposed into two components: temporal search and answer prediction, which can be formulated as:
\begin{equation}
    P_{\theta}(A, C \mid \tilde{V}, Q) = \underbrace{P_{\theta}(C \mid \tilde{V}, Q)}_{\emph{Temporal Search}} \cdot \underbrace{P_{\theta}(A \mid C, \tilde{V}, Q)}_{\emph{Answer Prediction}}.
    \label{eq:reasoning_decomposition}
\end{equation}

\paragraph{Dynamic Video Frames.} During the interleaved thinking process, the model autonomously explores the video by searching for additional clips. At reasoning step $k$, if the model outputs a \texttt{search} instruction, it is also required to specify the temporal boundaries $t_s^k$ and $t_e^k$ to be explored, along with a corresponding textual query $q^k$. The video environment then executes a frame retrieval function to obtain additional $F$ frames $V_k = \texttt{search}(V;\, t_s^k, t_e^k, q^k, F) = \{f_k^1, f_k^2, \ldots, f_k^{F}\}$. This function serves as an interface to the policy model $\pi_{\theta}$, employing a small VLM (e.g., SigLIP~\citep{zhai2023siglip}) to calculate the similarity among frames within the specified temporal clip $[t_s^k, t_e^k]$, as well as the relevance with the textual query $q^k$. The most informative $F$ frames are then sampled using determinantal point process (DPP)~\citep{kulesza2012determinantal}, which has been widely used for information retrieval~\citep{chen2018fast-map-dpp, celis2018p-dpp, sun2025mdp3}. This operation significantly improves the efficiency of temporal search, and more details can be found in Section~\ref{sec:CFM}.

\subsection{GRPO with Completeness Self-Verification}
\label{sec:grpo-csv}

\begin{figure}[t]
    \centering
    \includegraphics[width=1.0\textwidth]{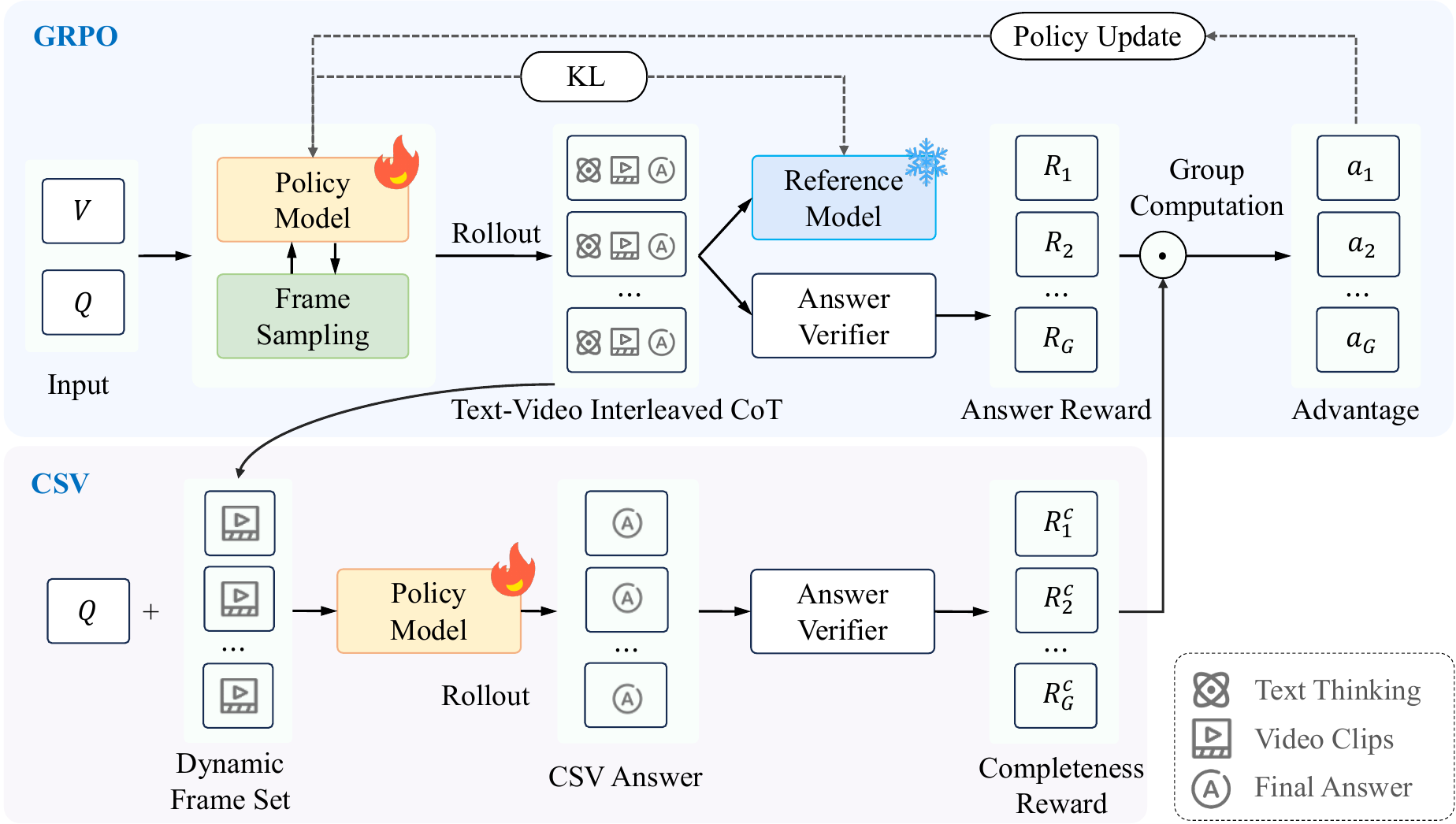}
    \caption{\textbf{Overall pipeline of GRPO-CSV.} Building upon the original GRPO, CSV extracts a dynamic frame set from the multi-modal CoT and constructs a vision-only CoT for re-answering. This design verifies that the searched dynamic frames provide sufficient evidence for correct reasoning, ensuring completeness and consistency without requiring explicit frame-level supervision.}
    \label{fig:grpo-csv}
    \vspace{-5mm}
\end{figure}

Evaluating temporal search typically requires frame-level annotations~\citep{ye2025rethinking}, which are time-consuming and labor-intensive. To circumvent this challenge, previous works~\citep{yu2025framevoyager, sun2025mdp3} have treated downstream video understanding task as a surrogate metric for assessing the searched frame set. However, these approaches are limited to selecting an optimal subset from a pre-defined pool of candidate frames, lacking interaction with and exploration of the video environment. Inspired by this, we design a \textbf{Completeness Self-Verification (CSV)} mechanism for GRPO, which is annotation-free and can be seamlessly integrated into RL training, serving as a complementary to the original outcome reward. The overall pipeline of GRPO-CSV is illustrated in Figure~\ref{fig:grpo-csv}.


\paragraph{GRPO-CSV.} We introduce CSV as a complement to GRPO with only outcome rewards. During the GRPO rollout phase, the policy model $\pi_{\theta}$ generates a text-video interleaved CoT $C$ and a final answer $A$. Applying rewards only to the final answer may reduce the effectiveness of intermediate search processes. To address this, we extract the video clips from $C$ to form a dynamic frame set $V_c$ as the input for the CSV phase. In the CSV rollout phase, the same policy model is required to re-answer the question $Q$ using only $V_c$, yielding a CSV answer $A_c$. Critically, the model is prohibited from any further temporal searching and must rely solely on the currently searched frames to answer the question. The CSV answer $A_c$ is expected to remain consistent with the original answer $A$:
\begin{equation}
    P_{\theta}(A_c \mid V_{c}, Q) \approx P_{\theta} (A \mid C, \tilde{V}, Q).
    \label{eq:csv_prob}
\end{equation}


\paragraph{Completeness Reward.} We design a completeness reward for the CSV phase, which is computed using the original answer $A$, the CSV answer $A_c$, and the ground-truth answer $A^*$ as follows:
\begin{equation}
    R_{c} = \mathbbm{1}[\textrm{Acc}(A, A^*) > 0.5] \cdot \textrm{Acc}(A_c, A^*).
    \label{eq:completeness_reward}
\end{equation}
where $\textrm{Acc}(A, A^*)$ and $\textrm{Acc}(A_c, A^*)$ denote the correctness scores of the original answer and the CSV answer, respectively, and $\mathbbm{1}[\cdot]$ is an indicator function activated only when the original answer $A$ is correct. This conditional design ensures that the CSV reward is applied only to promising reasoning trajectories, encouraging meaningful temporal search while verifying both the sufficiency of acquired visual evidence and the consistency between the reasoning process and the final answer.


\paragraph{Format Reward.} The format reward enforces adherence to a predefined schema throughout the multi-turn reasoning process, validating the structural integrity of the entire trajectory rather than individual steps. During reasoning, each step must follow either the \texttt{<think>...</think><tool\_call>...</tool\_call>} pattern for temporal search or the \texttt{<think>...</think><answer>...</answer>} pattern for the final response. We assign a binary score to the full trajectory: 1 if all steps are structurally valid, and 0 otherwise.


\paragraph{Accuracy Reward.} We evaluate answer accuracy for two task types. For multiple-choice questions, we extract the option letter from the model’s output and perform an exact match with the ground-truth option. For open-ended questions, we adopt an LLM-as-a-Judge approach~\citep{ZhengC00WZL0LXZ23LLM-as-a-Judge} to assess the semantic agreement between the model’s final answer and the reference answer. The scores for both cases are given in binary form, with 1 indicating alignment with the standard answer.


\paragraph{Overall Reward.} The total reward is the sum of completeness, format, and accuracy components:
\begin{equation}
    R = R_{c} + R_{\text{fmt}} + R_{\text{acc}}.
    \label{eq:overall_reward}
\end{equation}
This composition encourages sufficient temporal exploration ($R_{c}$), consistent reasoning structures ($R_{\text{fmt}}$), and correct final answers ($R_{\text{acc}}$), enhancing the model’s ability to understand long-form videos.


\subsection{Model Training}


\paragraph{Dataset Construction.} A fundamental challenge in RL for long-video reasoning lies in the fact that a large number of samples in the existing datasets can be solved through pure linguistic bias, reducing the reliance on temporal search. Moreover, some noisy samples remain unsolvable even under ideal temporal search, preventing the model from effectively exploring the video. To address these challenges, we implement a two-stage data filtering pipeline to construct a high-quality dataset tailored to video reasoning. In the first stage, we remove samples that the policy model can solve correctly using only 4 uniformly sampled frames, thereby discouraging reliance on linguistic shortcuts. In the second stage, we further discard samples that remain unsolvable even with multiple temporal searches and numerous video frames, ensuring active video exploration. Additional details of this filtering pipeline are provided in Section~\ref{sec:dataset_construction_details}. And we enhance dataset diversity through the incorporation of samples sourced from Haystack-Ego4D~\citep{ye2025rethinking}, VideoMarathon~\citep{lin2025unleashing_videomarathon}, and CinePile~\citep{cinepile}. A detailed analysis of the dataset is presented in Section~\ref{sec:dataset_analysis}.


\paragraph{Model Training.} We employ a two-stage training scheme for our TimeSearch-R. In the first stage, supervised fine-tuning (SFT) serves as a cold start, guiding the model to follow the correct reasoning format and enabling effective policy learning in the subsequent RL stage. SFT training adopts the above dataset construction pipeline, using GPT-4o~\citep{GPT4o} to generate the text-video interleaved reasoning processes and the corresponding final answers. Following practices in the text domain~\citep{jin2025search_r1}, we mask the temporal search results during training to force the model to learn meaningful temporal windows and textual queries. The objective in this stage is to minimize the standard cross-entropy loss over reasoning tokens, while excluding masked video tokens from gradient computation. Building on this cold-start, we further conduct RL post-training based on the proposed GRPO-CSV algorithm to stimulate the temporal reasoning capability of the model.


\section{Experiments}
\label{sec:experiments}

\input{tables/lvhaystack.tex}

\subsection{Experimental Setup}

\paragraph{Baselines.} To comprehensively evaluate the effectiveness of TimeSearch-R, we compare it against three types of baselines: (1) Advanced foundation models with static frame sampling, including both API models~\citep{GPT4o,team2024gemini} and open-source models~\citep{bai2025qwen25vltechnicalreport}. (2) State-of-the-art temporal search agents, such as VideoAgent~\citep{wang2024videoagent}, T*~\citep{ye2025rethinking}, and VideoTree~\citep{wang2025videotree}. (3) Video reasoning models like Video-R1~\citep{feng2025videor1}.

\paragraph{Datasets.} We evaluate TimeSearch-R on two tasks: (1) Temporal search on Haystack-LVBench and Haystack-Ego4D~\citep{ye2025rethinking}, where the task is modeled as long video needle-in-a-haystack, measuring temporal and visual similarity as well as QA accuracy. (2) Long-form video understanding on VideoMME~\citep{videomme}, MLVU~\citep{MLVU}, and LongVideoBench~\citep{wu2024longvideobench}.

\paragraph{Evaluation Metrics.} Besides the original metrics used in the benchmarks, we additionally introduce two metrics to assess the quality of the text-video interleaved thinking process for ablation study. Among them, \textit{completeness} measures whether the searched frame set is sufficient for the correct answer, while \textit{consistency} measures the alignment between intermediate reasoning and the final answer. Further details on these two metrics are provided in Section~\ref{sec:completeness_consistency_metrics}.

\paragraph{Implementation Details.} We train TimeSearch-R based on Qwen2.5-VL-7B-Instruct~\citep{bai2025qwen25vltechnicalreport}. In the RL training, we use the AdamW~\citep{loshchilov2017adamw} optimizer with a learning rate of 1e-6, a KL penalty coefficient $\beta$ = 0.005, and a batch size of 4 with 8 rollouts per prompt. We limit each search operation to retrieving at most 8 frames from a specified temporal clip, with up to 8 search steps in total. Training is conducted on 32 A100 GPUs. See more details in Section~\ref{sec:training_details}.

\input{tables/longvideobench.tex}

\subsection{Main Results}

\paragraph{Temporal Search.} On the temporal search task, TimeSearch-R establishes a new state-of-the-art on LV-Haystack, as shown in Table~\ref{tab:temporal_search}. Under a budget of 8 keyframes, our method achieves an $F_1$ score of 8.1 in temporal similarity, more than three times the previous best result of 2.5 obtained by T*. In visual similarity, TimeSearch-R reaches an $F_1$ score of 69.2, surpassing the previous SOTA method VideoAgent by 5.5, and even outperforming the retrieval-based method and T* with larger keyframe budgets. For the needle-in-a-haystack QA, our TimeSearch-R consistently outperforms the advanced API model GPT-4o, achieving 52.1\% accuracy on Haystack-LVBench and 53.5\% on Haystack-Ego4D. These results demonstrate the superiority of end-to-end learned temporal search strategies over handcrafted workflows based on human heuristics.


\paragraph{Long-Form Video Understanding.} Our TimeSearch-R also achieves strong performance on the long-form video understanding task, which is shown in Table~\ref{tab:long_video_understanding}. On VideoMME, our method reaches an overall accuracy of 66.6\%, surpassing the base model Qwen2.5-VL by 1.5\%. As the duration of the video increases, our method can achieve more gains, from 0.5\% on short videos to 1.4\% on long videos, demonstrating that temporal search becomes more valuable when the video length increases. On MLVU and LongVideoBench, TimeSearch-R achieves 71.5\% and 60.1\%, improving over the base model by 1.3\% and 4.1\%, respectively. Compared with video search agents, TimeSearch-R outperforms VideoAgent and T* on VideoMME by 10.6\% and 2.5\%, highlighting the advantage of end-to-end
optimization. Notably, our method consistently surpasses the latest video reasoning model Video-R1 across all benchmarks, validating that text-video interleaved reasoning is more effective than text-only reasoning for long-form video understanding. Moreover, directly applying temporal search to Qwen2.5-VL through CoT prompting without additional training actually degrades performance, underscoring the necessity of RL post-training with the proposed GRPO-CSV.


\subsection{Ablation Studies}
\label{sec:ablation_studies}

\begin{figure}[t]
    \centering
    \begin{minipage}{0.75\textwidth}
        \centering
        \input{tables/ablation_grpo_csv.tex}
        \vspace{4mm}
        \subcaption{Ablation Results}
        \label{fig:ablation_grpo_csv}
    \end{minipage}
    \hfill
    \begin{minipage}{0.24\textwidth}
        \centering
        \includegraphics[width=\textwidth]{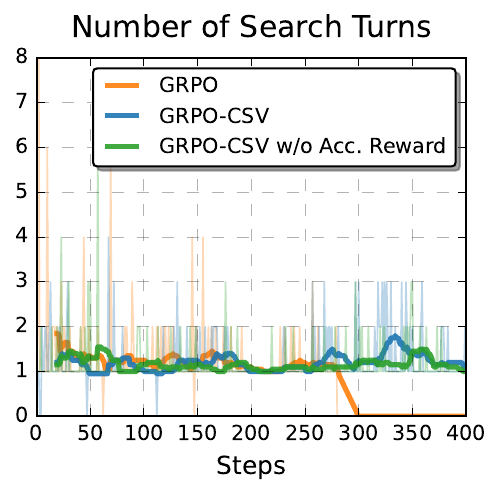}
        \vspace{-6mm}
        \subcaption{Training Dynamics}
        \label{fig:training_dynamics}
    \end{minipage}
    \vspace{-2mm}
    \caption{\textbf{Ablation study of GRPO-CSV.} (a) Comparison of different training schemes on temporal search and long-form video understanding. (b) When CSV is removed, training begins to collapse. The model gradually reduces the number of search calls and eventually stops searching altogether.}
    \label{fig:ablation_training_table}
    \vspace{-3mm}
\end{figure}

\paragraph{Training Scheme.} We explore the impact of different training stages in Table~\ref{fig:ablation_grpo_csv}, from zero-shot CoT to SFT and finally RL, yielding two key findings: (1) \textbf{SFT enables search capability:} The model cannot perform the search well only through zero-shot CoT prompts. SFT allows the model to rapidly acquire temporal search skills, dramatically improving temporal $F_1$ from 0.0 to 7.8 and searched frame completeness from 44.2\% to 60.5\%. (2) \textbf{RL enhances video reasoning:} While RL provides modest improvements to temporal similarity and search completeness, its primary advantage lies in boosting overall understanding performance. The post-training stage improves reasoning consistency by 2.6\%, which in turn raises QA accuracy from 59.2\% to 66.6\%.


\paragraph{GRPO-CSV Component.} We further conduct an ablation study on the components of GRPO-CSV in Figure~\ref{fig:ablation_training_table}, and obtain three key findings: (1) \textbf{GRPO reduces search completeness.} Without CSV as a complement, GRPO drops completeness from 60.5\% to 57.2\% and temporal $F_1$ from 7.8 to 7.4, demonstrating that outcome-only rewards lead to insufficient temporal exploration. (2) \textbf{GRPO-CSV improves training stability.} As illustrated in Figure\ref{fig:training_dynamics}, removing CSV causes training to collapse around step 300, after which the model ceases to make search calls and completeness drops to zero. (3) \textbf{GRPO-CSV with accuracy reward achieves the best QA performance.} While completeness reward alone achieves the highest completeness and consistency, it slightly reduces QA accuracy by 0.3\%. Combining GRPO-CSV with accuracy reward leads to the best overall QA performance.


\input{tables/ablation_dataset.tex}

\paragraph{Data Composition.} We also analyze the data composition in RL training, as shown in Table~\ref{tab:ablation_dataset}, revealing the contributions of data filtering and domain diversity. Without data filtering, RL training leads to a substantial performance drop compared to the original Qwen2.5-VL. This degradation arises because linguistic biases induce zero advantage in GRPO group computation: when questions can be trivially answered through linguistic shortcuts, all rollouts achieve perfect accuracy and completeness, yielding no learning signal and severely hindering RL efficiency and training stability. After applying data filtering, the model trained solely on egocentric data recovers baseline performance, but the lack of diversity weakens the benefits of RL. By incorporating exocentric data to enhance domain diversity, the model achieves its best general QA accuracy of 66.6\%. Notably, although the training data only includes general long-video QA tasks, RL training significantly boosts the model’s temporal and action reasoning capabilities, improving them by 7.4\% and 5.7\%, respectively. This remarkable performance demonstrates that TimeSearch-R learns fundamental cognitive patterns through end-to-end policy optimization, validating the strong generalization of our proposed GRPO-CSV algorithm.


\subsection{Case Studies}
\label{sec:case_studies}

\begin{figure}[t]
    \centering
    \includegraphics[width=1.0\textwidth]{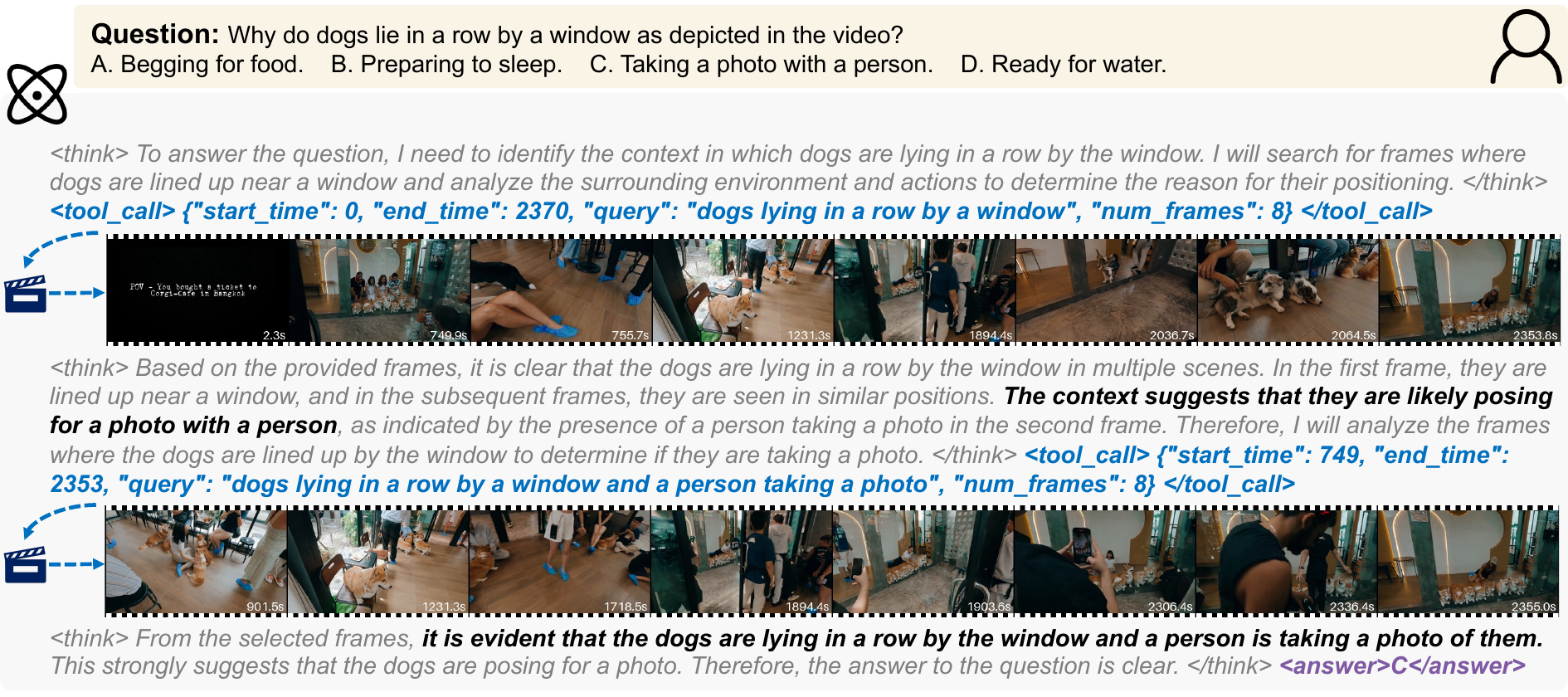}
    \vspace{-3mm}
    \caption{\textbf{Hypothesis-driven search.} Given the context that dogs are lying in a row across multiple scenes and remain still, the model hypothesizes that they are waiting to be photographed. It then searches for the person taking a photo to gather supporting evidence and provides the final answer.}
    \label{fig:hypothesis_driven_search}
    \vspace{-3mm}
\end{figure}

We analyze the search patterns that emerge during end-to-end RL training, demonstrating how the model executes temporal search within its reasoning process in a manner analogous to human cognition. These search patterns exhibit adaptability and flexibility across different task types:

\textbf{Hypothesis-driven search.} The model formulates hypotheses based on limited context and executes targeted searches to gather additional video frames as supporting evidence. (Figure~\ref{fig:hypothesis_driven_search})

\textbf{Confirmation or elimination.} When the initially sampled dynamic frame set provides insufficient support for an answer, the model employs multi-faceted search strategies or elimination methods to collect additional evidence and reduces uncertainties. (Figure~\ref{fig:case_search_confirmation} and ~\ref{fig:case_elimination_method})
 
\textbf{Sequential search.} The model performs segment-by-segment analysis to accomplish temporal reasoning tasks that require understanding sequential relationships. (Figure~\ref{fig:case_sequential_search})

\section{Related Work}
\label{sec:related_work}


\paragraph{Temporal Search for Long-Video Understanding.}
Traditional video understanding methods rely on static frame sampling, such as uniform sampling or heuristic-based  strategies~\citep{li2024llavaonevision,chen2024internvl,bai2025qwen25vltechnicalreport}, which fail to adapt to varying information density and evolving reasoning contexts.
Recent work has explored more sophisticated mechanisms.
Similarity-based methods like KeyVideoLLM~\citep{liang2024keyvideollm} achieve significant compression while maintaining performance , while learning-based approaches such as Frame-Voyager~\citep{yu2025framevoyager} rank frame combinations based on prediction losses, emphasizing task-specific selection.
Advanced semantic frameworks have emerged to address temporal dependencies.
Logic-in-Frames~\citep{guo2025logicinframes} defines logical relations including spatial co-occurrence and temporal proximity to guide dynamic frame sampling.
T*~\citep{ye2025rethinking} reframes temporal search as spatial search with adaptive zooming mechanisms.
Interactive agents like VideoAgent~\citep{wang2024videoagent} and VideoTree
~\citep{wang2025videotree} enable multi-turn temporal exploration through prompt-driven orchestration.
However, none of the aforementioned methods adopt end-to-end optimization, resulting in suboptimal search strategies.

\paragraph{Reinforcement Learning for Multimodal Reasoning.}
Recent advances have explored RL to enhance reasoning capabilities in LLMs.
GRPO~\citep{deepseekai2025deepseekr1incentivizingreasoningcapability} demonstrates that outcome-based rewards can effectively elicit complex reasoning.
Search-R1~\citep{jin2025search_r1} extends this paradigm to text-based search tasks, showing that RL can facilitate adaptive information retrieval.
Approaches like MM-Eureka~\citep{meng2025mmeurekaexploringfrontiersmultimodal} and LMM-R1~\citep{peng2025lmmr1empowering3blmms} have successfully applied RL to enhance multimodal reasoning, but focus primarily on static image understanding rather than dynamic video interaction.
Video-R1~\citep{feng2025videor1} applies GRPO to video reasoning but limits the thinking process to prue text without visual interaction, while DeepEyes~\citep{deepeyes} uses RL for high-resolution image understanding through adaptive cropping operations but focuses on spatial rather than temporal exploration.
Despite these advances, applying RL to interactive long video understanding remains largely unexplored and presents unique challenges.

\section{Conclusion}

In this work, we propose TimeSearch-R, a framework that reformulates temporal search as text-video interleaved thinking to learn optimal search strategies directly from data. To enhance temporal search through RL, we propose CSV as a complement to the outcome-only reward of GRPO, addressing the challenges of insufficient temporal exploration and inconsistent logical reasoning. TimeSearch-R achieves strong performance on both temporal search and long-form video understanding tasks, while exhibiting distinct search patterns across different task types. We hope this work contributes meaningful progress toward advancing long video understanding powered by reinforcement learning.

\clearpage
\bibliography{iclr2026_conference}
\bibliographystyle{iclr2026_conference}

\clearpage

\appendix

\begin{center}
  \LARGE
  \justifying{\textsc{TimeSearch-R: Adaptive Temporal Search for Long-Form Video Understanding via Self-Verification Reinforcement Learning}}
\end{center}
\begin{center}
  \Large
  \vspace{0.5em} \textsc{Appendix} \\
  \vspace{0.5em}
\end{center}

This appendix provides more details about our methods, dataset, training, more case studies, broader impacts, as well as the LLM usage, organized as follows:
\begin{itemize}
    \item Section~\ref{sec:CFM}: Search Function
    \item Section~\ref{sec:dataset_details}: Dataset Details
    \item Section~\ref{sec:prompt_design}: Prompt Design
    \item Section~\ref{sec:completeness_consistency_metrics}: Evaluation Metrics
    \item Section~\ref{sec:efficiency_analysis}: Efficiency Analysis
    \item Section~\ref{sec:training_details}: Training Details
    \item Section~\ref{sec:more_case_studies}: More Case Studies
    \item Section~\ref{sec:broader_impact}: Boarder Impacts
    \item Section~\ref{sec:llm_usage}: LLM Usage
\end{itemize}

\section{Search Function}
\label{sec:CFM}


\subsection{Frame Selection}

The video \texttt{search} function selects the most informative frames within predicted temporal clips.
Specifically, we leverage determinantal point process (DPP)~\citep{kulesza2012determinantal} as the search optimization for its ability to naturally balance query relevance and diversity that penalizes redundancy, which has been widely applied in information retrieval~\citep{celis2018p-dpp, sun2025mdp3}.

Recall the definition of \texttt{search} in Sec.~\ref{sec:task_formulation}, it aims to select $F$ optimal frames guided by a temporal clip $[t_s, t_e]$ and a query $q$ from the original video $V$.
First, the function first subsamples $N$ candidate frames $\mathcal{F}_{[t_s,t_e]} = \{v_i\}_{i=1}^{N}$ within the temporal clip. 
Subsequently, we obtain a visual embedding $\mathbf{h}_i \in \mathbb{R}^d$ for each candidate frame in $\mathcal{F}_{[t_s,t_e]}$, and a query embedding $\mathbf{q} \in \mathbb{R}^d$ for $q$.
Then we define the pairwise cosine similarity for candidate frames as $S_{ij} = \mathbf{h}_i^\top \mathbf{h}_j$ and compute an unnormalized query relevance score for each frame as $\tilde r_i = \mathbf{q}^\top \mathbf{h}_i$, which is rescaled to $[0,1]$ by min-max normalization $r_i = \frac{\tilde r_i - \min \, \tilde{\mathbf{r}}}{\max \, \tilde{\mathbf{r}} - \min \, \tilde{\mathbf{r}} + \epsilon}$,
where $\epsilon$ is a small constant to avoid division by zero.
The kernel is constructed by diagonal conditioning with these relevance weights:
\begin{equation}
\tilde{\mathbf{L}} = \mathrm{diag}(\mathbf{r})\, \mathbf{S}\, \mathrm{diag}(\mathbf{r}),
\end{equation}
which is equivalent to $\tilde L_{ij} = r_i r_j \, \mathbf{h}_i^\top \mathbf{h}_j$.
The optimal subset $V^* \subset \mathcal{F}_{[t_s,t_e]}$ with $|V^*| = F$ is then obtained through fast greedy MAP inference~\citep{chen2018fast-map-dpp}:
\begin{equation}
V^* = \arg\max_{S \subseteq \mathcal{F}_{[t_s,t_e]}, |S|=F} \det(\tilde{\mathbf{L}}_S).
\end{equation}
This formulation ensures that selected frames are both diverse and relevant to the query.
When available frames are fewer than $F$, the search function degrades to uniform temporal sampling.

\subsection{Frame Representation}

The selected clip frames are sparse and non-uniform.
To maintain the temporal pace, we attach an explicit absolute timestamp to each frame by inserting a short text token with the time in seconds (e.g., ``12.3s'') immediately before the image.
This simple interleaving of timestamp text and the corresponding image maintains absolute temporal grounding when inter-frame intervals vary and complements the native temporal ids.
Explicit absolute timestamp augmented frame representation has also been observed to improve temporal capability in prior work on long-video temporal grounding~\citep{pan2025timesearch}.
For uniformly sampled preview frames, we employ the native dynamic-FPS and absolute time encoding following Qwen2.5-VL~\citep{bai2025qwen25vltechnicalreport}, which bind image token sequences to temporal ids aligned with real absolute timestamps.

\section{Dataset Details}
\label{sec:dataset_details}

\subsection{Dataset Construction}
\label{sec:dataset_construction_details}

\begin{figure}[t]
    \centering
    \includegraphics[width=1.0\textwidth]{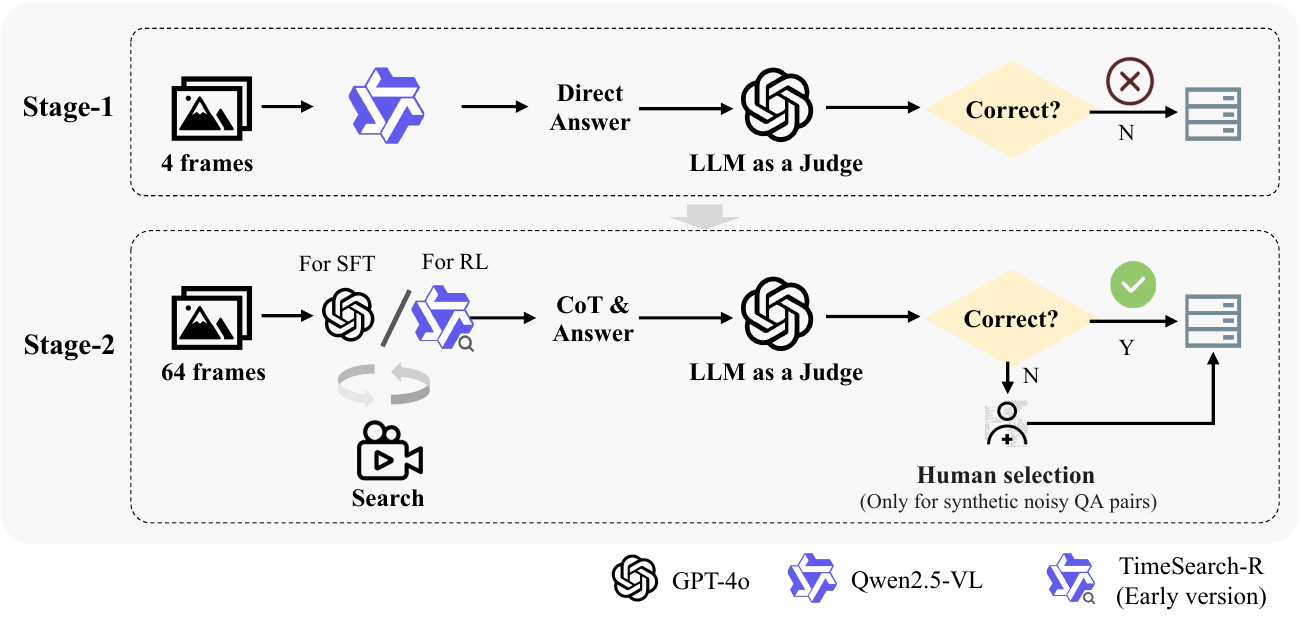}
    \caption{\textbf{Illustration of the proposed two-stage data filtering pipeline.}}
    \label{fig:data_filtering}
\end{figure}

To ensure high-quality training data, we implement a two-stage filtering pipeline as shown in Fig.~\ref{fig:data_filtering}.

\textbf{Stage 1: Visual Dependency Filtering.}
We uniformly sample 4 frames from each video and feed them along with the question to Qwen2.5-VL for inference.
Questions that can be correctly answered with this limited visual information are considered to have low visual dependency and are subsequently filtered out.
Only questions requiring richer visual context proceed to the next stage.

\textbf{Stage 2: Search Usefulness Filtering.}
We increase the frame input to up to 64 frames and employ different LVLMs to perform dynamic temporal search for question-relevant video segments.
Specifically, we use GPT-4o to generate SFT (Supervised Fine-Tuning) data and an early version of TimeSearch to obtain RL (Reinforcement Learning) training data.
Although this stage produces CoT, only the CoT generated by GPT-4o is used for SFT training, while RL training utilizes only the question-answer pairs.
To avoid search format errors, we implement format validation for LVLMs' responses, automatically retrying the model until obtaining properly formatted answers.

\textbf{Human Selection for VideoMarathon (Panda-70M).}
Given that VideoMarathon's training set contains automatically generated question-answer pairs with potential unanswerable questions or incorrect ground-truth answers, we conduct manual annotation to ensure data quality.
To minimize annotator bias in model answer evaluation, we establish a structured annotation protocol.
First, annotators assess question reasonableness based on video content, filtering out unanswerable or ambiguous questions.
Subsequently, annotators provide manual answers and compare them with synthetic ground-truth labels, removing data samples that are inconsistent with human responses.

\subsection{Dataset Analysis}
\label{sec:dataset_analysis}

\begin{figure}[t]
    \centering
    \includegraphics[width=1.0\textwidth]{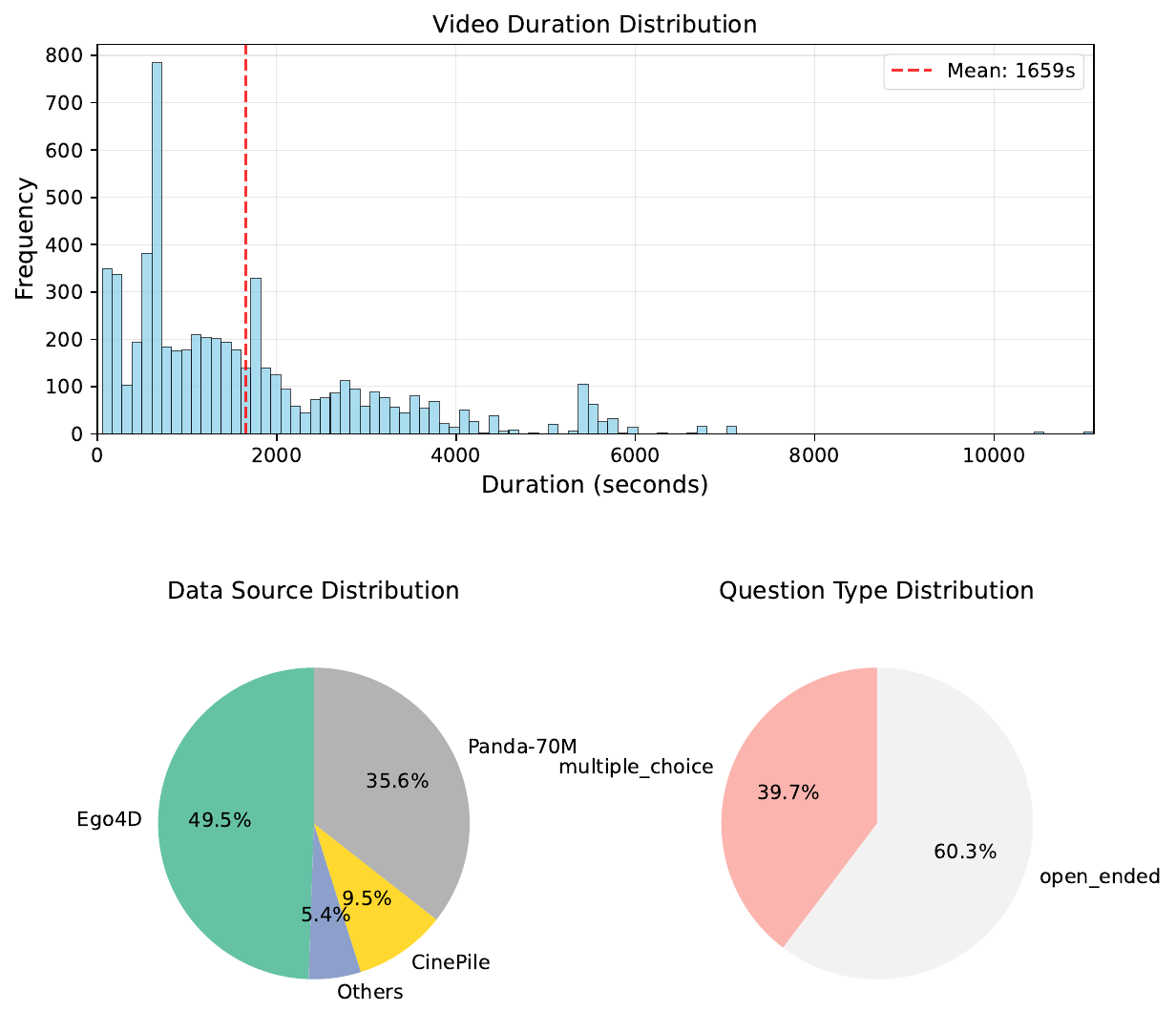}
    \caption{\textbf{Dataset analysis.} (1) The training set is mainly composed of long videos. The average length is 1659 seconds, and the maximum length exceeds 10,000 seconds. (2) Egocentric QA pairs come from Haystack-Ego4D, and Exocentric QA data mainly from VideoMarathon and Cinepile, where VideoMarathon employs Panda-70M as the video source. (3) Question types include multiple-choice and open-ended questions. To obtain open-ended QA pairs, we convert some multiple-choice tasks into open-ended questions.
    }
    \label{fig:dataset_overview}
\end{figure}

The dataset exhibits a pronounced long-tail distribution in video duration with a mean length of 1,659 seconds.
Most videos are shorter than 2,000 seconds, while a nontrivial tail extends beyond one hour, posing significant challenges for static frame sampling.
This distribution motivates adaptive temporal search and multi-turn interaction to progressively retrieve evidence under tight keyframe budgets.

We curate data from four major sources to ensure coverage of diverse visual domains and camera styles. 
As shown in Fig.~\ref{fig:dataset_overview}, Ego4D from Haystack-Ego4D~\citep{ye2025rethinking} training set contributes 49.5\% of samples, providing egocentric daily activities with frequent viewpoint changes.
Panda-70M from VideoMarathon~\citep{lin2025unleashing_videomarathon} accounts for 35.6\%, expanding the variety of internet videos with heterogeneous motion patterns and scene dynamics.
CinePile~\citep{cinepile} provides 9.5\% of short videos with narrative structure and rapid scene transitions.
The remaining 5.4\% are from other sources and serve to reduce distributional bias.

Question types are intentionally imbalanced toward open-ended reasoning to better evaluate generative capabilities.
Open-ended questions make up 60.3\% of the data and emphasize step-by-step analysis, temporal grounding, and explanation quality.
Multiple-choice questions comprise 39.7\% and offer reliable automatic evaluation signals that complement outcome rewards in RL.

This composition yields wide coverage over motion intensity, scene diversity, and narrative structure while maintaining sufficient automatic evaluability.
The mixture of long-tail durations and open-ended questions creates a setting where end-to-end RL and adaptive temporal search offer clear benefits over single-shot heuristics.

\section{Prompt Design}
\label{sec:prompt_design}

We design prompts to standardize interaction formats, minimize ambiguity, and provide explicit priors for temporal reasoning.
Fig.~\ref{fig:system_prompt}--\ref{fig:csv_prompt} show the templates used during training and evaluation.

\paragraph{System Prompt.}
We follow the tool-use specification of the base Qwen2.5-VL family~\citep{bai2025qwen25vltechnicalreport} and adopt its \texttt{tool\_call} schema for invoking temporal search.
This design ensures deterministic parsing by the environment and stable credit assignment for RL, as illustrated in Fig.~\ref{fig:system_prompt}.

\begin{promptbox}{System Prompt}
\begin{lstlisting}[style=mystyle]
You are a helpful video assistant.
# Tools
You may call one or more functions to assist with the user query.
You are provided with function signatures within <tools></tools> XML tags:
<tools>
{"type": "function", "function": {"name": "seek_video_frames", "description": "Search and select video frames according to textual query and temporal window. Time is in seconds.", "parameters": {"type": "object", "properties": {"query": {"type": "string", "description": "The query is used to describe the object, scene, or event of interest in the video thoroughly and clearly. "}, "start_time": {"type": "number", "description": "Start time of the segment of interest. "}, "end_time": {"type": "number", "description": "End time of the segment of interest. "}, "num_frames": {"type": "integer", "description": "Number of frames to sample (maximum 8). Default is 8."}}, "required": ["query"]}}}
</tools>
For each function call, return a json object with function name and arguments within <tool_call></tool_call> XML tags:
<tool_call>
{"name": <function-name>, "arguments": <args-json-object>}
</tool_call>
\end{lstlisting}
    \end{promptbox}
\tcbcaption{\textbf{The system prompt with tools.} \label{fig:system_prompt}}

\paragraph{Question Answering Prompt.}
The QA template enforces thorough reasoning inside \texttt{<think>} before any tool call or final answer.
It restricts the output to exactly one of two formats and allows at most eight rounds of \texttt{<tool\_call>}.
It explicitly provides the line \textit{"The video duration: \{duration\} seconds."} to help the model produce absolute timestamps better.
See Fig.~\ref{fig:qa_prompt}.

\begin{promptbox}{Question Answering}
\begin{lstlisting}[style=mystyle]
You must ALWAYS conduct thorough reasoning inside <think> and </think> tags BEFORE calling any tool or answering the question.
You must invoke tools to explore any video content you are interested in within <tool_call> </tool_call> tags. 
You are allowed to use <tool_call></tool_call> tags for a maximum of 8 rounds.
When you have enough information to answer the question, provide your answer within <answer> </answer> tags. Your answer should be supported by evidence from the video.
Your output must follow the format: <think>Your reasoning process</think><tool_call>Parameters</tool_call> or <think>Your reasoning process</think><answer>Your answer</answer>Question: {question}
The video duration: {duration} seconds.
\end{lstlisting}
    \end{promptbox}
\tcbcaption{\textbf{The template for question answering.} \label{fig:qa_prompt}}

\paragraph{Clip Frame Sampling and Search Response.}
After a search, the template returns the selected frames and their corresponding timestamps.
If the frames are sufficient, the model must place the final answer in \texttt{<answer>}.
Otherwise, the template asks the model to call the tool again with different parameters in JSON, thereby encouraging reflection and re-query.
See Fig.~\ref{fig:cfm_prompt}.

\begin{promptbox}{Temporal Search Response}
\begin{lstlisting}[style=mystyle]
Here are selected frames. They are located at {timestamps}.
If the frames provided above are sufficient to answer the user's question, please put your final answer within <answer></answer>.
Otherwise invoke the tool again with different parameters in JSON format.
\end{lstlisting}
    \end{promptbox}
\tcbcaption{\textbf{The response template of the temporal search.} \label{fig:cfm_prompt}}

\paragraph{Completeness Self-Verification Prompt.}
The CSV template asks the model to answer as briefly as possible and to say \textit{"I don't know"} when the visual evidence is insufficient.
No tools are available in this stage, which prevents new searches and ensures the answer is grounded only on the dynamic frame set gathered earlier.
See Fig.~\ref{fig:csv_prompt}.

\begin{promptbox}{Completeness Self-Verification}
\begin{lstlisting}[style=mystyle]
You are a helpful assistant. Please answer visual questions as briefly as possible. When you don't have enough visual information, please say 'I don't know'.
\end{lstlisting}
    \end{promptbox}
\tcbcaption{\textbf{The template for CSV reasoning.} \label{fig:csv_prompt}}

\section{Evaluation Metrics}
\label{sec:completeness_consistency_metrics}

\paragraph{Completeness Rate.}
We measure the proportion of cases where the dynamic visual context alone suffices to produce the correct answer.
Concretely, after the multi-turn search, we re-answer the question using only the gathered dynamic frame set and disallow further search, following the CSV procedure in Sec.~\ref{sec:grpo-csv} and the prompt illustrated in Fig.~\ref{fig:csv_prompt}.
The resulting correctness is computed with the same task-specific accuracy used elsewhere, averaged over the whole dataset.

\paragraph{Consistency Rate.}
Consistency evaluates whether the intermediate reasoning coherently supports the final answer under the given question.
We prompt a LLM model (GPT-4o) with the question, the reasoning text extracted from \texttt{<think>...</think>}, and the final answer from \texttt{<answer>...</answer>}, using the format in Fig.~\ref{fig:consistency_prompt} that requires a structured output: a short analysis in \texttt{<think>} followed by \texttt{<answer>} equal to ``Yes'' or ``No''.
In implementation, we parse the LLM's output to obtain the binary decision; ``Yes'' is counted as 1 and ``No'' as 0, and any parsing failure is treated as 0.
The Consistency Rate is the dataset average of these binary outcomes.

\begin{promptbox}{Consistency Score Evaluation}
\begin{lstlisting}[style=mystyle]
<system prompt>
You are a careful and logical reviewer. Your task is to verify whether the given reasoning process and the final answer are consistent in addressing the given question.

Please carefully read the following information:

Question: <Question>
Reasoning Process: <Reasoning>
Final Answer: <Answer>

Please follow this format strictly:
<think> Your analysis here </think> <answer> Yes/No </answer>
\end{lstlisting}
    \end{promptbox}
\tcbcaption{\textbf{The template for calculating consistency.} \label{fig:consistency_prompt}}


\section{Efficiency Analysis}
\label{sec:efficiency_analysis}
\input{tables/efficiency.tex}

TimeSearch-R attains an end-to-end latency of 13.4 seconds on the Haystack-Ego4D test set, yielding a 61.6\% speed-up over the 34.9-second latency of VideoAgent.
Despite T$^*$ operating with the lightweight YOLO-World-110M detector and completing inference in 11.1 seconds, our method maintains a comparable runtime while avoiding the complexity of hand-crafted scheduling.
As shown in Tab.~\ref{tab:temporal_search}, TimeSearch-R markedly surpasses these baselines in temporal search metrics and QA accuracy, underscoring the effectiveness of reinforcement-driven temporal policies.

\section{Training Details}
\label{sec:training_details}

\input{tables/training_hyperparameters.tex}

We summarize the key hyperparameters in Table~\ref{tab:training_hyperparameters} for reproducibility.

\paragraph{Training Configuration.}
TimeSearch-R employs a distributed training setup using PyTorch's native distributed data parallel framework with ZeRO-3 memory optimization through DeepSpeed.
The training process leverages gradient accumulation to simulate larger batch sizes while maintaining memory efficiency on GPU clusters.
We utilize mixed precision training with bfloat16 to accelerate computation while preserving numerical stability, coupled with Flash Attention 2.0 for efficient attention computation.

\paragraph{GRPO Training Setup.}
The reinforcement learning phase uses Group Relative Policy Optimization with 8 generations per prompt to provide sufficient policy gradient estimates.
The KL divergence penalty coefficient $\beta$ is set to 0.005 to balance between reward optimization and policy regularization.
We employ VLLM in colocate mode for efficient inference during rollout generation, enabling faster policy updates.
This RL training stage is implemented on top of the TRL library~\citep{vonwerra2020trl}, following standard practice for outcome-driven policy optimization in large language models.

\paragraph{Video Processing Configuration.}
The model processes videos with a maximum of 768 frames and allocates up to 10,240 tokens for video content representation.
Each interaction turn is limited to 8 search operations, with a maximum of 8 interaction turns per question to ensure comprehensive temporal exploration while maintaining computational efficiency.
Frame tokens are dynamically allocated between 12 and 256 tokens per frame based on content complexity and relevance.

\section{More Case Studies}
\label{sec:more_case_studies}
This section provides more case studies of TimeSearch-R, including successful cases and failed cases.

\begin{figure*}[t]
    \centering
    \includegraphics[width=1.0\textwidth]{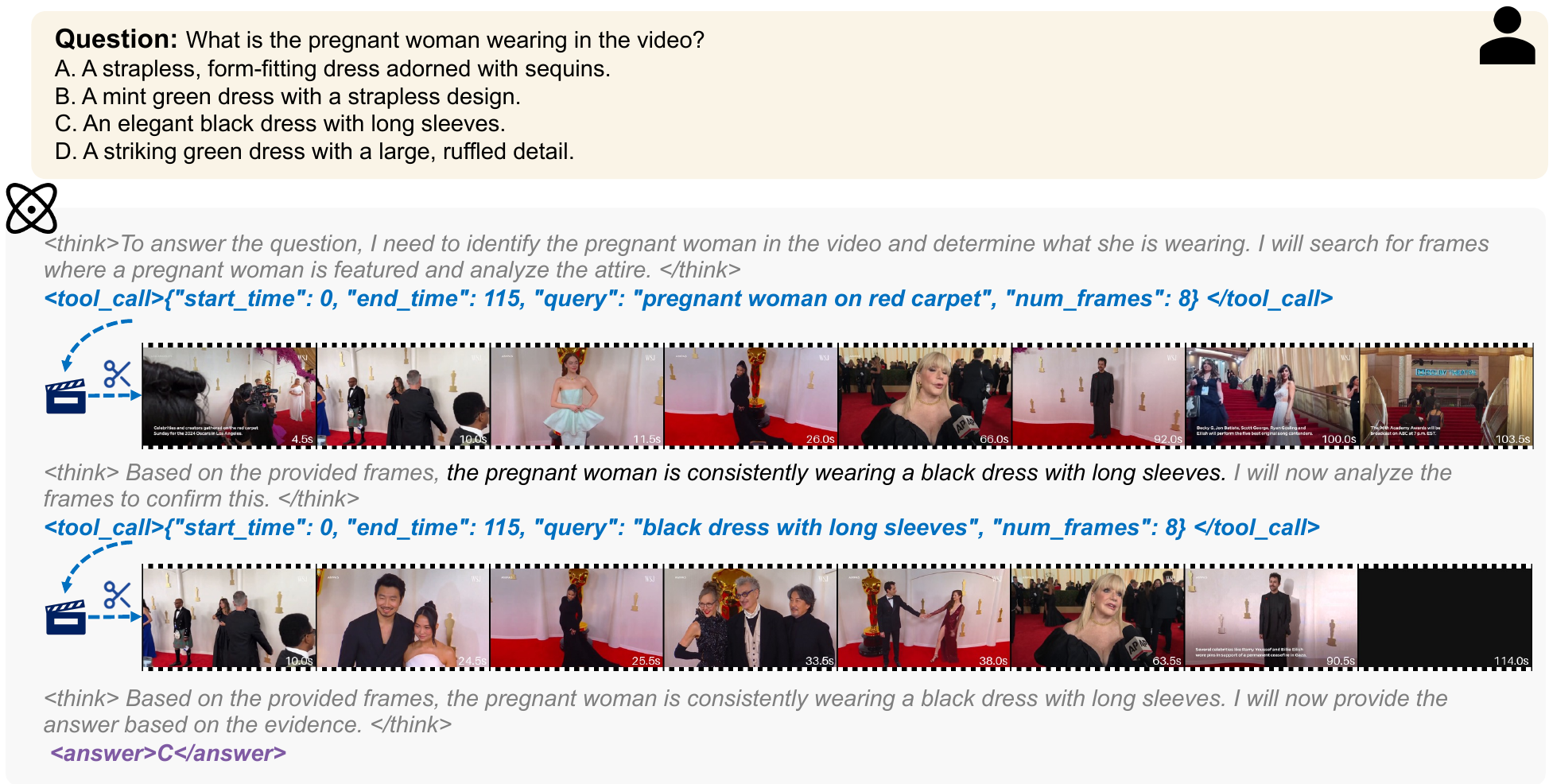}
    \caption{\textbf{Search pattern: search confirmation.}}
    \label{fig:case_search_confirmation}
\end{figure*}

\begin{figure*}[t]
    \centering
    \includegraphics[width=1.0\textwidth]{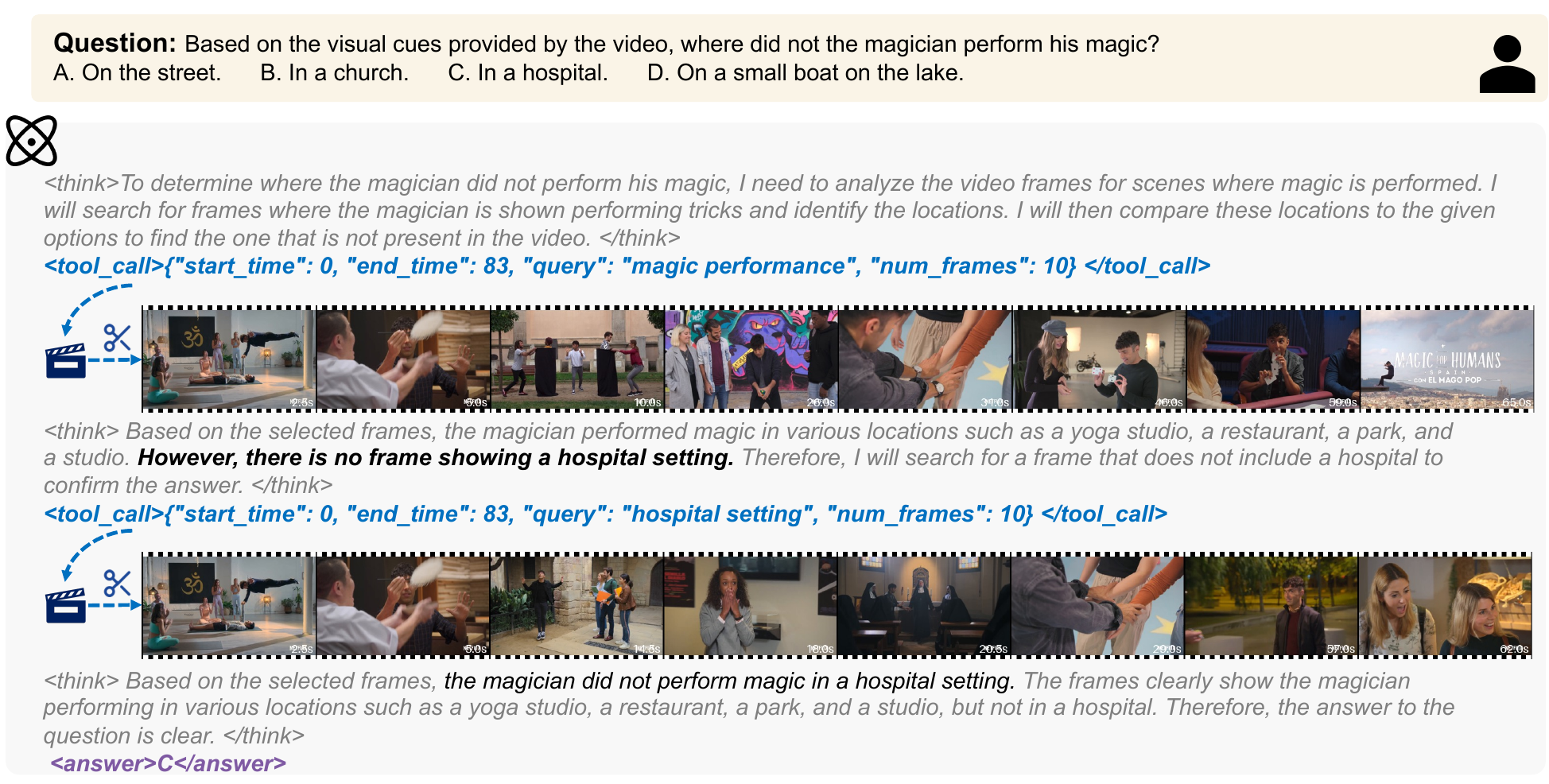}
    \caption{\textbf{Search pattern: elimination method.}}
    \label{fig:case_elimination_method}
\end{figure*}

\begin{figure*}[t]
    \centering
    \includegraphics[width=1.0\textwidth]{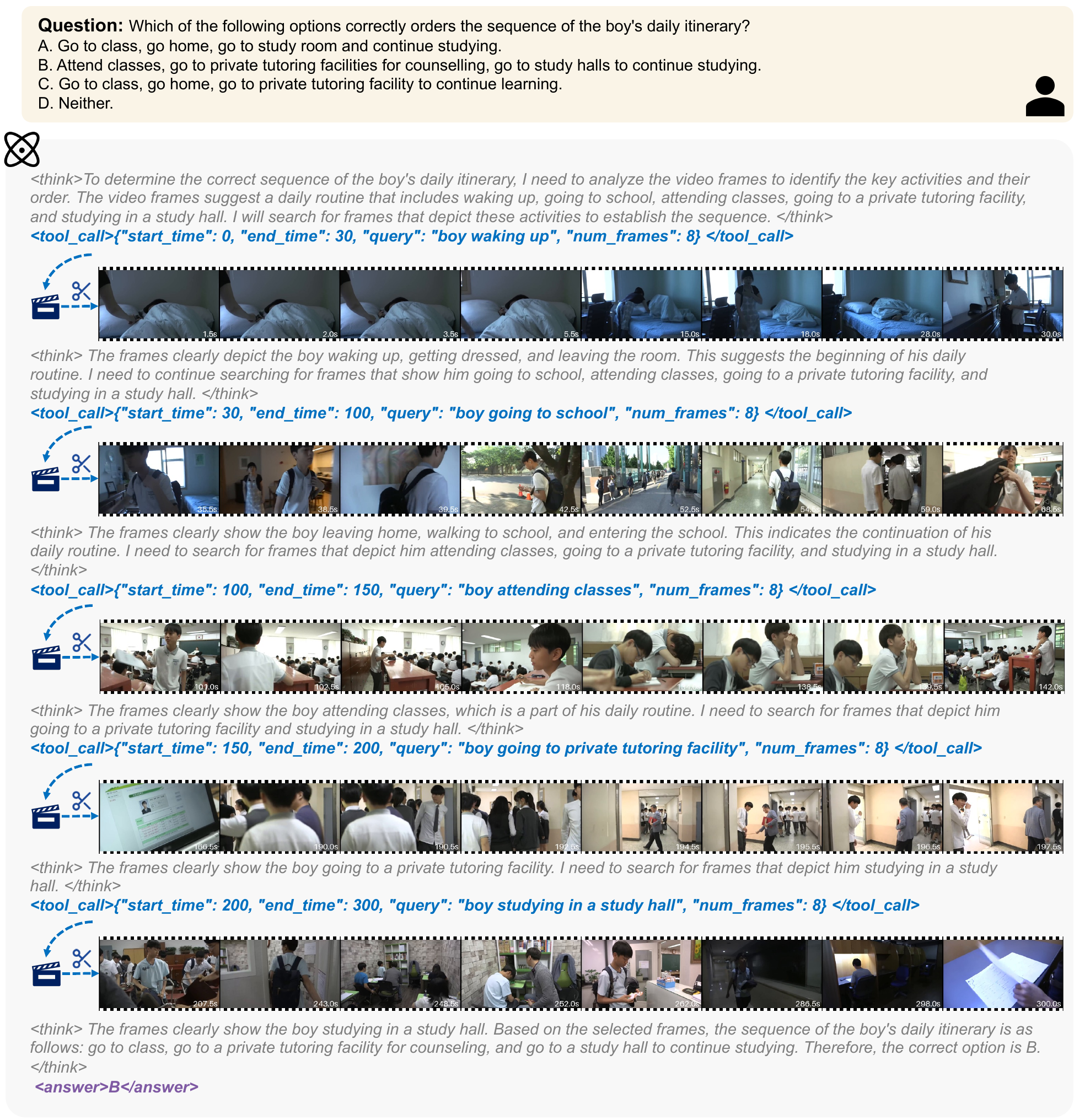}
    \caption{\textbf{Search pattern: sequential search.}}
    \label{fig:case_sequential_search}
\end{figure*}

\paragraph{Successful Cases.}
These representative success cases illustrate how TimeSearch-R conducts multi-turn exploration to accumulate decisive visual evidence while maintaining alignment between the reasoning trace and the final answer.
They encompass confirmation (Fig.~\ref{fig:case_search_confirmation}), elimination (Fig.~\ref{fig:case_elimination_method}), and sequential exploration patterns (Fig.~\ref{fig:case_sequential_search}), collectively demonstrating that the policy preserves the high completeness and consistency reported in Sec.~\ref{sec:completeness_consistency_metrics}.

\begin{figure*}[t]
    \centering
    \includegraphics[width=1.0\textwidth]{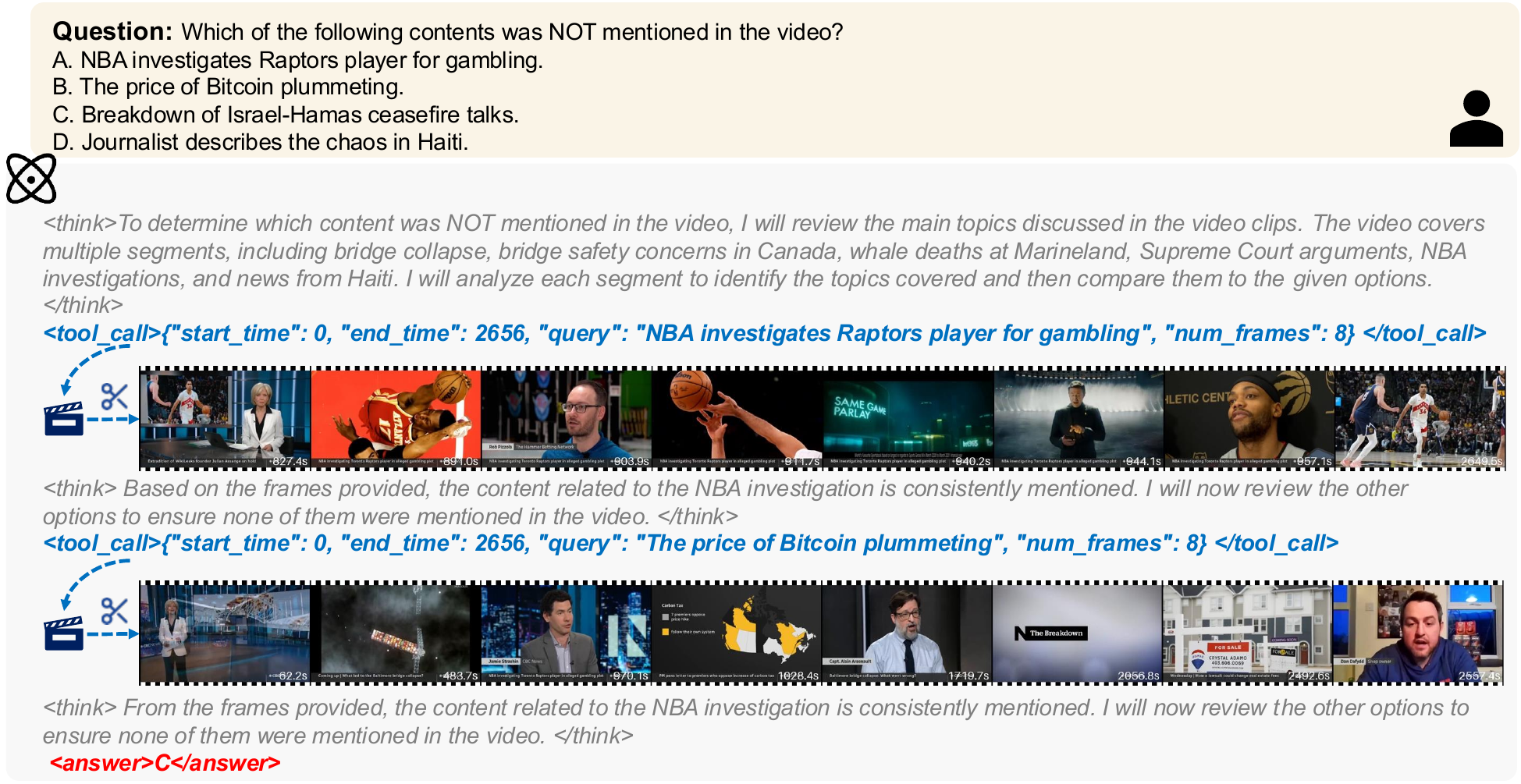}
    \caption{\textbf{Failure case: insufficient search.} There were 4 options in total, but only 2 were reviewed before the search was terminated.}
    \label{fig:case_early_stop}
\end{figure*}

\begin{figure*}[t]
    \centering
    \includegraphics[width=1.0\textwidth]{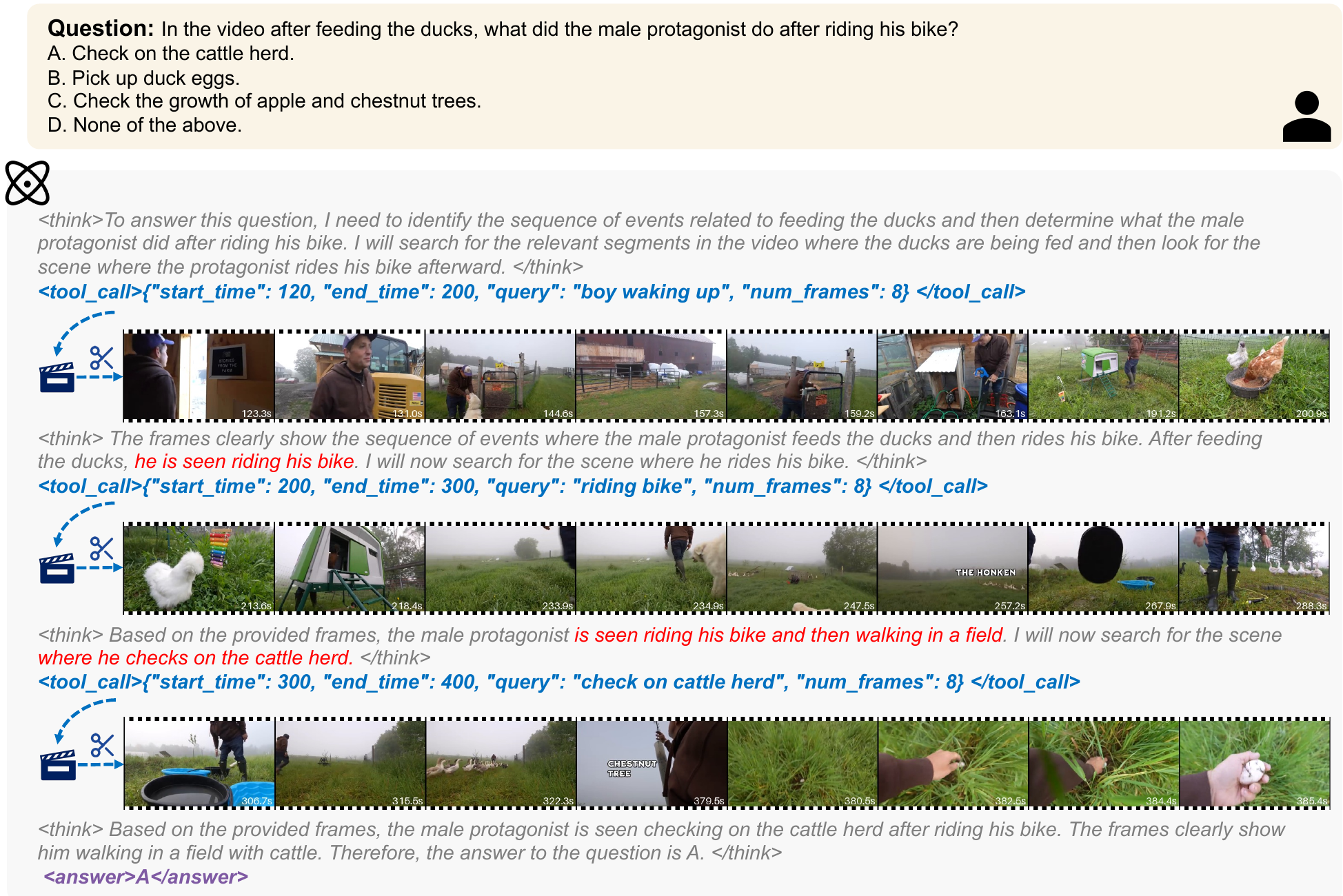}
    \caption{\textbf{Failure case: visual hallucination.} No information related to riding was found in the search results.}
    \label{fig:case_hallucination}
\end{figure*}

\paragraph{Failed Cases.}
Figure~\ref{fig:case_early_stop} illustrates a residual failure where the policy halts after reviewing only two of four candidate segments, leading to an incorrect answer.
Figure~\ref{fig:case_hallucination} illustrates a failure where the model hallucinates information related to riding.

\section{Broader Impacts}
\label{sec:broader_impact}

TimeSearch-R contributes to several important areas beyond the immediate technical contributions:

\paragraph{Advancing Video Interpretability and Explainability.}
TimeSearch-R introduces interleaved text-video reasoning traces that provide transparent insights into the model's decision-making process.
The completeness and consistency criteria we propose enable quantitative assessment of long-form video explanations, making temporal search decisions auditable and interpretable.
This advancement represents a significant step toward more explainable AI systems in the video domain, where understanding the reasoning process is crucial for building trust and ensuring reliability.

\paragraph{Transforming Video Reasoning from Static to Dynamic Paradigms.}
Our approach fundamentally shifts the paradigm from static frame sampling to dynamic, interactive reasoning in video understanding.
By operationalizing hypothesis-driven exploration through iterative temporal search, we promote a new methodology that emphasizes transparent, stepwise evidence gathering.
This contrasts sharply with traditional one-shot inference over fixed visual contexts, encouraging researchers to develop more adaptive and interactive AI systems.
The demonstrated effectiveness of our approach may inspire broader adoption of similar interactive paradigms across various multimodal tasks.

\paragraph{Exploring Scalable Weakly-Supervised Process Rewards.}
We introduce outcome-based process supervision that eliminates the need for costly process annotations.
Through the integration of weak supervision and reinforcement learning via completeness self-verification, our method successfully aligns intermediate search decisions with correct outcomes.
This approach offers a scalable solution for training complex interactive systems without requiring fine-grained procedural labels, potentially reducing annotation costs and enabling broader application across diverse domains.

\section{The Use of Large Language Models}
\label{sec:llm_usage}

The authors declare that Large Language Models were used in this paper for polishing the writing. Specifically, the LLM assisted with tasks such as grammar checking, sentence simplification, and improving the overall fluency of the text. It is important to note that the LLM was not used for any literature review or research ideation. All research ideas and experimental analyses presented in the paper were solely conducted by the authors.

\clearpage

\end{document}

%% file: math_commands.tex
\usepackage{amsmath,amsfonts,bm}

\def\eqref#1{equation~\ref{#1}}

\def\1{\bm{1}}

\DeclareMathAlphabet{\mathsfit}{\encodingdefault}{\sfdefault}{m}{sl}
\SetMathAlphabet{\mathsfit}{bold}{\encodingdefault}{\sfdefault}{bx}{n}



%% file: marcos.tex
\definecolor{codegreen}{HTML}{478058}
\definecolor{codegray}{rgb}{0.5,0.5,0.5}
\definecolor{codepurple}{HTML}{4F5E80} 
\definecolor{backcolour}{rgb}{0.95,0.95,0.92}
\lstdefinestyle{mystyle}{
    backgroundcolor=\color{backcolour},
    commentstyle=\color{codegreen},
    keywordstyle=\color{magenta},
    numberstyle=\tiny\color{codegray},
    stringstyle=\color{codepurple},
    basicstyle=\ttfamily\scriptsize,
    breakatwhitespace=false,
    breaklines=true,
    captionpos=b,
    keepspaces=true,
    frame=none,
    numbersep=5pt,
    showspaces=false,
    showstringspaces=false,
    showtabs=false,
    tabsize=2,
    breakindent=0pt,
    postbreak=\mbox{},
    breakautoindent=false
}

\newtcolorbox{promptbox}[2][]{
    enhanced, 
    breakable,
    center title,
    left*=0pt, right*=0pt,
    boxsep=2pt, left=5pt, right=5pt,
    skin first=enhanced,
    skin middle=enhanced,
    skin last=enhanced,
    colback  = backcolour,
    fonttitle=\bfseries\rmfamily,
    fontupper=\scriptsize,
    title={\footnotesize\strut{#2}},
    #1
}

\newcommand{\tcbcaption}[1]{
\noindent\begin{minipage}{1.0\textwidth}
\captionof{figure}{#1}
\end{minipage}
}

%% file: tables/lvhaystack.tex
\begin{table}[t]
    \centering
    \caption{\textbf{Temporal search performance.} We report temporal similarity, visual similarity, and question-answering (QA) accuracy on Haystack-LVBench, as well as QA accuracy on Haystack-Ego4D test-tiny subset. Baseline results are directly cited from \citet{ye2025rethinking}. $^\dagger$ indicates the average number of keyframes determined by the model adaptively.}
    \label{tab:temporal_search}
    \resizebox{1.0\textwidth}{!}{
        \begin{tabular}{l|c|c|ccc|ccc|cc}
            \toprule 
            \multirow{2}[2]{*}{\textbf{Method}} & \multirow{2}[2]{*}{\textbf{Base Model}} & \multirow{2}[2]{*}{\textbf{\# Frame}} & \multicolumn{3}{c|}{\textbf{Temporal}} & \multicolumn{3}{c|}{\textbf{Visual}} & \multicolumn{2}{c}{\textbf{QA}} \\
            \cmidrule(lr){4-6} \cmidrule(lr){7-9} \cmidrule(lr){10-11}
            &  &  & P & R & $F_1$ & P & R & $F_1$ & LVBench & Ego4D \\
            \midrule
            \rowcolor{gray!10}\multicolumn{11}{c}{\textit{Static Frame Sampling}} \\ 
            Uniform  & Qwen2.5VL-7B & 8  & 1.4 & 6.3 & 2.2 & 56.0 & 72.0 & 62.7 & 33.7 & 32.0 \\
            Uniform  & GPT-4o & 8  & 1.4 & 6.3 & 2.2 & 56.0 & 72.0 & 62.7 & 47.1 & 41.5 \\
            \textcolor{gray}{Uniform} & \textcolor{gray}{GPT-4o} & \textcolor{gray}{32} & \textcolor{gray}{1.4} & \textcolor{gray}{24.9} & \textcolor{gray}{2.7} & \textcolor{gray}{58.7} & \textcolor{gray}{81.6} & \textcolor{gray}{67.3} & \textcolor{gray}{50.5}  & \textcolor{gray}{45.5} \\
            \midrule
            \rowcolor{gray!10}\multicolumn{11}{c}{\textit{Adaptive Temporal Search}} \\
            VideoAgent~\citep{wang2024videoagent} & GPT-4  & 10.1$^\dagger$ & 1.2 & \underline{8.5} & 2.1 & 58.8 & \underline{73.2} & \underline{64.7} & - & -  \\
            Retrieval-based~\citep{ye2025rethinking} & GPT-4o & 8  & 1.5 & 6.3 & 2.3 & \underline{63.1} & 65.5 & 64.1 & - &  - \\
            T*~\citep{ye2025rethinking} & GPT-4o & 8 & \underline{1.6} & 7.1 & \underline{2.5} & 58.4 & 72.7 & 64.3 & \underline{51.9} & \underline{45.0} \\
            \textcolor{gray}{Retrieval-based~\citep{ye2025rethinking}} & \textcolor{gray}{GPT-4o} & \textcolor{gray}{32}  & \textcolor{gray}{1.3} & \textcolor{gray}{21.8} & \textcolor{gray}{2.4} & \textcolor{gray}{59.9} & \textcolor{gray}{80.8} & \textcolor{gray}{67.8}  & \textcolor{gray}{-} &  \textcolor{gray}{-} \\
            \textcolor{gray}{T*~\citep{ye2025rethinking}} & \textcolor{gray}{GPT-4o} & \textcolor{gray}{32}  & \textcolor{gray}{1.7} & \textcolor{gray}{28.2} & \textcolor{gray}{3.1} & \textcolor{gray}{58.3} &  \textcolor{gray}{83.2} & \textcolor{gray}{67.8} & \textcolor{gray}{53.1} & \textcolor{gray}{46.5} \\
            \midrule
            \rowcolor{gray!10}\multicolumn{11}{c}{\textit{Text-Video Interleaved Reasoning}} \\
            TimeSearch-R (Ours) & Qwen2.5VL-7B & 8.8$^\dagger$ & \textbf{5.4} &  \textbf{22.3} & \textbf{8.1} & \textbf{63.2} & \textbf{76.4} & \textbf{69.2} & \textbf{52.1} & \textbf{53.5} \\
            \bottomrule
        \end{tabular}
    }
\end{table}

%% file: tables/longvideobench.tex
\begin{table}[t]
  \centering
  \caption{\textbf{Video understanding performance.} E2E stands for end-to-end optimization. \# Frame represents the number of input frames. $^\dagger$ indicates the keyframes produced by temporal search.}
  \label{tab:long_video_understanding}
  \vspace{-3mm}
  \resizebox{1.0\textwidth}{!}{
    \begin{tabular}{l|cc|cccc cc }
    \toprule  
    \multirow{2}[2]{*}{\textbf{Model}} & \multirow{2}[2]{*}{\textbf{E2E}} & \multirow{2}[2]{*}{\textbf{\# Frame}}& \multicolumn{4}{c}{\textbf{VideoMME (w/o sub)}} & \textbf{MLVU} & \textbf{LVB} \\ 
    \cmidrule(lr){4-7} \cmidrule(lr){8-8} \cmidrule(lr){9-9} & & & short & medium & long & \textbf{overall} & \textbf{m-avg} & \textbf{val} \\
    \midrule
    \rowcolor{gray!10}\multicolumn{9}{c}{\textit{Static Frame Sampling}} \\
    Qwen2.5VL-7B~\citep{bai2025qwen2} & \ding{51} & 768 & 76.3 & 66.0 & 54.6 & 65.1 & 70.2 & 56.0 \\
    GPT-4o~\citep{GPT4o} & \ding{51} & 384 & 80.0 & 70.3 & 65.3 & 71.9 & 64.6 & 66.7 \\
    Gemini-1.5-Pro~\citep{team2024gemini} & \ding{51} & 1 fps & 81.7 & 74.3 & 67.4 & 75.0 & -- & 64.0 \\
    \midrule
    \rowcolor{gray!10}\multicolumn{9}{c}{\textit{Adaptive Temporal Search}} \\
    VideoAgent (GPT-4)~\citep{wang2024videoagent} & \ding{55} & 87$^\dagger$ & --  & -- & 49.0 & 56.0 & -- & -- \\
    VideoTree (GPT-4)~\citep{wang2025videotree} & \ding{55} & 128$^\dagger$ & 67.8 & 59.9  & 54.2 & --  & -- & -- \\
    T$^*$ (GPT-4o)~\citep{ye2025rethinking} & \ding{55} & 32$^\dagger$ & 69.5 & 63.5 & 59.3 & 64.1 & -- & -- \\
    \midrule
    \rowcolor{gray!10}\multicolumn{9}{c}{\textit{Text-only Reasoning}} \\
    Video-R1-7B~\citep{feng2025videor1} & \ding{51} & 32 & 71.1 & 59.0 & 49.4 & 59.9 & 61.6 & 56.4 \\
    Video-R1-7B~\citep{feng2025videor1} & \ding{51} & 768 & 74.1 & 65.1 & 55.6 & 65.7 & 68.4 & 58.1 \\
    \midrule
    \rowcolor{gray!10}\multicolumn{9}{c}{\textit{Text-Video Interleaved Reasoning}} \\
    Qwen2.5VL-7B + Search & \ding{55} & 768 & 53.4 & 53.8 & 48.2 & 51.8 & 58.9 & 49.1 \\
    TimeSearch-R-7B (Ours) & \ding{51} & 768 & \textbf{76.8} & \textbf{67.1} & \textbf{56.0} & \textbf{66.6} & \textbf{71.5} & \textbf{60.1} \\
    $\Delta$ \textit{(v.s. Qwen2.5VL-7B)} & -- & -- & +0.5 & +1.1 & +1.4 & +1.5 & +1.3 & +4.1 \\
    \bottomrule
    \end{tabular}
  }
  \vspace{-4mm}
\end{table}

%% file: tables/ablation_grpo_csv.tex
\definecolor{ForestGreen}{RGB}{34,139,34}

\resizebox{\textwidth}{!}{
\begin{tabular}{@{}lcccccc@{}}
    \toprule
    \multirow{2}[2]{*}{\textbf{Method}} & \multicolumn{3}{c}{\textbf{Haystack-LVBench}} & \multicolumn{3}{c}{\textbf{VideoMME}} \\
    \cmidrule(lr){2-4} \cmidrule(lr){5-7}
    & P & R & $F_1$ & Comp. & Cons. & Acc. \\
    \midrule
    Qwen2.5-VL w/ search & 0.0 & 0.0 & 0.0 & 44.2 & 59.4 & 51.8 \\
    SFT & 7.4 & 11.6 & 7.8 & 60.5 & 69.2 & 59.2 \\
    \midrule
    \textcolor{orange}{GRPO (Before Collapse)} & 5.2\textcolor{red}{$_{-2.2}$} & 18.8\textcolor{ForestGreen}{$_{+7.2}$} & 7.4\textcolor{red}{$_{-0.4}$} & 57.2\textcolor{red}{$_{-3.3}$} & 69.3\textcolor{ForestGreen}{$_{+0.1}$} & 65.1\textcolor{ForestGreen}{$_{+5.9}$} \\
    \textcolor{ForestGreen}{GRPO-CSV w/o Acc. Rwd} & 6.1\textcolor{red}{$_{-1.3}$} & 19.8\textcolor{ForestGreen}{$_{+8.2}$} & 8.2\textcolor{ForestGreen}{$_{+0.4}$} & 61.2\textcolor{ForestGreen}{$_{+0.7}$} & 75.3\textcolor{ForestGreen}{$_{+6.1}$} & 64.8\textcolor{ForestGreen}{$_{+5.6}$} \\
    \textcolor{blue}{GRPO-CSV w/ Acc. Rwd} & 5.4\textcolor{red}{$_{-2.0}$} & 22.3\textcolor{ForestGreen}{$_{+10.7}$} & 8.1\textcolor{ForestGreen}{$_{+0.3}$} & 60.2\textcolor{red}{$_{-0.3}$} & 71.8\textcolor{ForestGreen}{$_{+2.6}$} & 66.6\textcolor{ForestGreen}{$_{+7.4}$} \\
   \bottomrule
\end{tabular}
}


%% file: tables/ablation_dataset.tex

\begin{table}[t]
    \centering
    \caption{\textbf{Ablation study of data composition.} Line 1 shows the accuracy of original Qwen2.5-VL.}
    \label{tab:ablation_dataset}
    \vspace{-2mm}
    \resizebox{\textwidth}{!}{
    \begin{tabular}{ ccc|cccc|cccc }
        \toprule
        \multirow{2}[2]{*}{\textbf{Ego}} & \multirow{2}[2]{*}{\textbf{Exo}} & \multirow{2}[2]{*}{\textbf{Filter}} & \multicolumn{4}{c}{\textbf{General}} & \multicolumn{4}{c}{\textbf{Reasoning}} \\
        \cmidrule(lr){4-9} \cmidrule(lr){8-11}
        & & & short & medium & long & overall & temporal & spatial & action & object \\
        \midrule
        -- & -- & -- & \underline{76.3} & \underline{66.0} & 54.6 & \underline{65.1} & 51.4 & \textbf{76.8} & 56.8 & \underline{59.5} \\
        \midrule
        \checkmark & \checkmark & & 74.2 & 62.7 & 51.3 & 62.8 & 40.1 & 67.9 & \underline{60.0} & 56.2 \\
        \checkmark & & \checkmark & 76.4 & 64.7 & \underline{54.9} & 65.3 & \underline{54.8} & 73.2 & 58.2 & 59.0 \\
        \checkmark & \checkmark & \checkmark & \textbf{76.8} & \textbf{67.1} & \textbf{56.0} & \textbf{66.6} & \textbf{58.8} & \underline{75.0} & \textbf{62.5} & \textbf{61.9} \\
        \bottomrule
    \end{tabular}
    }
    \vspace{-4mm}
\end{table}


%% file: tables/efficiency.tex
\begin{table}[h]
    \centering
    \caption{\textbf{Efficiency evaluation on Haystack-Ego4D.}
    Baseline results are directly cited from \citet{ye2025rethinking}.
    We report the overall latency of temporal search and answering.
    Evaluations are conducted on the Haystack-Ego4D using A100 GPUs.
    Temporal search metrics are reported in Tab.~\ref{tab:temporal_search}.}
    \label{tab:efficiency}
        \begin{tabular}{lccc}
        \toprule
        \textbf{Method} & \textbf{Question Grounding} & \textbf{Frame Retrieval} & \textbf{Latency (sec)} $\downarrow$ \\
        \midrule
        VideoAgent & GPT4 & CLIP-1B & 34.9 \\
        Retrieval-based & -- & YOLO-world-110M & 32.2 \\
        T$^*$ (Detector-based) & LLaVA-OV-7B & YOLO-world-110M & 11.1 \\
        \midrule
        TimeSearch-R & -- & SigLIP-400M & 13.4 \\
        \bottomrule
        \end{tabular}
\end{table}

%% file: tables/training_hyperparameters.tex
\begin{table}[th!]
\centering
\caption{\textbf{Training hyperparameters of TimeSearch-R.}}
\label{tab:training_hyperparameters}
\begin{tabular}{llr}
\toprule
\textbf{Category} & \textbf{Parameter} & \textbf{Value} \\
\midrule
\multirow{4}{*}{\textbf{Video Processing}} 
& Max FPS & 2 \\
& Max Frames per Video & 768 \\
& Total Video Tokens & 10,240 \\
& Min Tokens per Frame & 12 \\
& Max Tokens per Frame & 256 \\
\midrule
\multirow{3}{*}{\textbf{Interaction Settings}} 
& Max Search Turns & 8 \\
& Max Completion Length per Turn & 256 \\
\midrule
\multirow{6}{*}{\textbf{GRPO Training}} 
& Number of Generations & 8 \\
& KL Penalty Coefficient ($\beta$) & 0.005 \\
& Scale Rewards & false \\
& Batch Size per GPU & 1 \\
& Gradient Accumulation Steps & 2 \\
\midrule
\multirow{3}{*}{\textbf{Infrastructure}} 
& DeepSpeed Configuration & ZeRO-3 Offload \\
& VLLM Mode & colocate \\
& Replay Buffer & true \\
\bottomrule
\end{tabular}%
\end{table}